\documentclass[nohyperref]{article}

\usepackage{microtype}
\usepackage{graphicx}
\usepackage{booktabs} 

\usepackage{hyperref}

\usepackage[accepted]{icml2022}

\usepackage{amsmath}
\usepackage{amssymb}
\usepackage{mathtools}
\usepackage{amsthm}

\usepackage[capitalize,noabbrev]{cleveref}

\usepackage{url}
\usepackage{wrapfig}
\usepackage{listings}
\usepackage{multirow}

\usepackage{subcaption}
\usepackage{float}

\usepackage{tikz}
\usetikzlibrary{shapes,arrows}
\usetikzlibrary{patterns}
\usetikzlibrary{decorations.pathreplacing, decorations.pathmorphing, calligraphy}

\tikzstyle{rectblock} = [rectangle, draw, thick, align=center]
\tikzstyle{block} = [rectblock, rounded corners]
\tikzstyle{boundingbox} = [very thick, gray]
\tikzstyle{dashblock} = [rectangle, draw, thick, align=center, dashed]
\tikzstyle{conc} = [ellipse, draw, thick, dashed, align=center]
\tikzstyle{netnode} = [circle, draw, very thick, inner sep=0pt, minimum size=0.5cm]
\tikzstyle{opnode} = [circle, draw, thick, inner sep=0pt, minimum size=0.5cm, align=center]
\tikzstyle{relunode} = [rectangle, draw, very thick, inner sep=0pt, minimum size=0.5cm]
\tikzstyle{line} = [draw, very thick, -latex']
\tikzstyle{arrow} = [draw, ->, thick]
\tikzstyle{attention} = [arrow, bend right]
\tikzstyle{mapsto} = [draw, |->, thick]
\tikzstyle{annobrace} = [draw, thick, decorate, decoration={brace, mirror, amplitude=0.5em}]
\tikzstyle{annotext} = [pos=0.5, text width=1.5cm, anchor=north, yshift=-0.25em, align=center]

\definecolor{bpurp}{HTML}{984ea3}
\definecolor{bblue}{HTML}{377eb8}
\definecolor{bgreen}{HTML}{4daf4a}
\definecolor{borange}{HTML}{ff7f00}
\definecolor{bred}{HTML}{a50f15}

\definecolor{mygreen}{HTML}{148219}
\definecolor{mysalmon}{HTML}{b06d78} 

\theoremstyle{plain}

\theoremstyle{definition}

\theoremstyle{remark}

\usepackage[textsize=tiny]{todonotes}

\icmltitlerunning{Explanations support learning relational and causal structure}

\begin{document}

\twocolumn[
\icmltitle{Tell me why! Explanations support learning relational and causal structure}




\icmlsetsymbol{equal}{*}

\begin{icmlauthorlist}
\icmlauthor{Andrew K. Lampinen}{dm}
\icmlauthor{Nicholas A. Roy}{dm}
\icmlauthor{Ishita Dasgupta}{dm}
\icmlauthor{Stephanie C. Y. Chan}{dm}
\icmlauthor{Allison C. Tam}{dm}
\icmlauthor{James L. McClelland}{dm}
\icmlauthor{Chen Yan}{dm}
\icmlauthor{Adam Santoro}{dm}
\icmlauthor{Neil C. Rabinowitz}{dm}
\icmlauthor{Jane X. Wang}{dm}
\icmlauthor{Felix Hill}{dm}
\end{icmlauthorlist}

\icmlaffiliation{dm}{DeepMind, London, UK}

\icmlcorrespondingauthor{Andrew Lampinen}{lampinen@deepmind.com}

\icmlkeywords{Explanations, RL, Reinforcement Learning, Relations, Causal Learning}

\vskip 0.3in
]



\printAffiliationsAndNotice{}  

\everypar{\looseness=-1}
\linepenalty=500
\begin{abstract}
Inferring the abstract relational and causal structure of the world is a major challenge for reinforcement-learning (RL) agents. For humans, language---particularly in the form of \textit{explanations}---plays a considerable role in overcoming this challenge. Here, we show that language can play a similar role for deep RL agents in complex environments. While agents typically struggle to acquire relational and causal knowledge, augmenting their experience by training them to predict language descriptions and explanations can overcome these limitations. We show that language can help agents learn challenging relational tasks, and examine which aspects of language contribute to its benefits. We then show that explanations can help agents to infer not only relational but also causal structure. Language can shape the way that agents to generalize out-of-distribution from ambiguous, causally-confounded training, and explanations even allow agents to learn to perform experimental interventions to identify causal relationships. Our results suggest that language description and explanation may be powerful tools for improving agent learning and generalization.
\end{abstract}

It is often argued that machine learning models---and deep learning models in particular---lack the human proficiencies for forming abstractions and inferring relational or causal structure~\citep[e.g.][]{fodor1988connectionism,lake2017building,pearl2018theoretical,marcus2020next,ichien2021visual,holyoak2021emergence, puebla2021can,geirhos2020shortcut}. These limitations can make it hard to train models that generalize out-of-distribution, or that reason in human-like ways, particularly for reinforcement learning (RL) agents that receive high-bandwidth input from raw pixels and must learn to act in partially-observable environments. 

Human learning of abstract, relational, and causal structure benefits substantially from language, and particularly \textit{explanations}. Language helps us to identify structure in the world, and to structure our thinking \citep{edmiston2015makes, lupyan2016centrality,dove2020more}. 
\emph{Explanations}---language that provides explicit information about appropriate abstractions and causal structure~\citep{keil2000explanation,lombrozo2006structure}---are particularly useful. Explanations mitigate credit assignment problems, by linking a concrete situation to reusable abstractions~\citep{lombrozo2006structure,lombrozo2006functional}, which helps humans to learn efficiently, from otherwise underspecified examples \citep{ahn1992schema}. They also help us make comparisons and master relational and analogical reasoning \citep{gentner2008relational,lupyan2008taking, edwards2019explanation}.  And they selectively highlight generalizable causal factors in a situation, improving our causal inference \citep{lombrozo2006functional}. Even explaining to ourselves, without feedback, can improve our generalization \citep{chi1994eliciting,rittle2006promoting,williams2010role}, perhaps because explanations form abstractions that are easier to recall and generalize \citep[cf.][]{dasgupta2021memory}.

Indeed, there has been increasing interest in using language or explanations as a learning signal for machines (e.g.,\ \citealp{ross2017right,mu-etal-2020-shaping,camburu2018snli,schramowski2020making}; see related work). That is, rather than seeking explanations post-hoc, to help humans understand the system \citep[e.g.,][]{chen2018learning,topin2019generation,xie2020explainable}, these works use explanations to help the system understand the task \citep[cf.][]{santoro2021symbolic}. Most machine learning from explanations focuses on supervised learning\footnote{The few studies on explanations in RL \citep{guan2020widening,tulli2020learning} do not explore relations or causal interventions.}, but explanations may be even more relevant to reinforcement learners. While supervised learners are theoretically limited~\citep[e.g.,][]{pearl2018theoretical}, RL agents can intervene and can therefore acquire causal knowledge~\citep[e.g.,][]{dasgupta2019causal,rezende2020causally}. At the same time, RL agents struggle with credit-assignment, abstraction, and generalization \citep{ghosh2021generalization,kirk2021survey}---the exact settings where explanations support humans. 
Both of these observations motivate exploring whether language explanations could help RL agents infer relational and causal structure in complex environments.  

\textbf{What is a language explanation?} We define a language explanation to be a string indicating a relationship between a situation, the agent's behavior, and abstract task structure. For example, after turning on an oven, an agent might receive an explanation like ``turning on the oven heats it up, to prepare for baking.'' This explanation conveys the abstract causal links between an action and a desired goal. By contrast, many true statements would not qualify as explanations, because they ignore the agent's behavior, task-relevant structure, or both; e.g., noting that ``the oven is silver'' would not be an explanation unless that fact is relevant to the task. We use the term ``explanation'' in this work to refer to a broad class of language utterances, including descriptions if they convey task-relevant abstractions; see Discussion.

We explore the benefits of explanations using tasks situated in rich 2D and 3D environments. To study relational learning and abstraction, we use a challenging relational tasks involving uniqueness---identifying the object from a set that is the \emph{odd-one-out} along one of multiple varying dimensions. To study causality, we first study learning in ambiguous, causally-confounded tasks, where multiple distinct features perfectly predict reward in training, but the features are dissociated in evaluation. We then study causal interventions, where agents must learn to perform experimental interventions to identify the causal structure of a particular episode.

In all settings, we find that deep RL performs poorly, but that learning and generalization improve substantially when agents learn to predict language explanations. Explanations help agents learn more effectively, preventing them from fixating on easy-but-inadequate ``shortcut'' features \citep{geirhos2020shortcut,hermann2020shapes}. Explanations also help agents to disentangle confounded features, and can shape the way that agents generalize out-of-distribution on deconfounded evaluations. They also enable agents to learn to perform experiments in order to identify the causal structure of each episode.  

To understand these effects better, we explore how different aspects of explanations contribute to their benefits We demonstrate that the most effective way to exploit explanations is to train agents to predict them, rather than simply observe them (prediction also avoids a need for explanations in evaluation). We further show that explanation prediction is learned much more rapidly than the tasks, supporting the idea that language helps agents learn task-relevant abstractions, which in turn make learning the task easier. We furthermore show that explanations that provide feedback relevant to the specific behavior of the agent are more effective than behavior-agnostic signals, even unsupervised auxiliary objectives (input reconstruction) that provide much more information. Thus, the distinct benefits of explanations can outperform or complement more generic auxiliary learning.

Taken together, these results suggest that generating explanations could be a powerful tool for augmenting RL in challenging tasks. Furthermore, explanations posed in natural language may be simpler for humans to produce than other forms of supervision \citep[e.g.,][]{cabi2019scaling,guan2020widening}. Thus, training agents to generate such explanations is a viable path towards both improved learning and generalization, and perhaps toward more human-like and interpretable agent behavior.

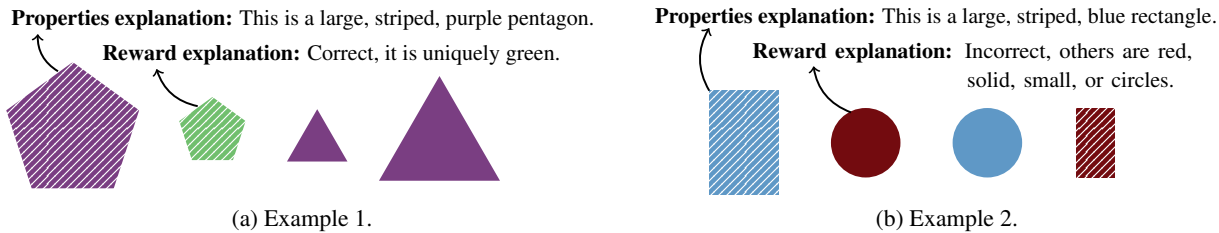
\begin{figure*}[thb]
\vskip-0.5em
\begin{subfigure}[b]{0.5\textwidth}
\centering
\resizebox{0.95\textwidth}{!}{
\begin{tikzpicture}
\node[fill=bpurp!80!black, regular polygon, regular polygon sides=5, minimum height=2cm] at (0, 0) (lp) {};
\node[fill=bgreen!80!white, regular polygon, regular polygon sides=5, minimum height=1cm] at (2, 0) (sp) {};
\draw[draw=white, pattern=north east lines, pattern color=white] (-1,-1) rectangle (2.75,1);
\node[fill=bpurp!80!black,regular polygon, regular polygon sides=3,  minimum height=1cm] at (3.5,-0.175)  {};
\node[fill=bpurp!80!black,regular polygon, regular polygon sides=3,  minimum height=2cm] at (5.25,-0.2) (lt) {};

\node[align=left, anchor=west] at (-1, 1.6) (exp_lp) {\small \textbf{Properties explanation:} This is a large, striped, purple pentagon.};
\draw[arrow, bend left] ([xshift=-6, yshift=-4]lp.north) to (-0.5, 1.4);

\node[align=left, anchor=west] at (0.3, 1.1) (exp_sp) {\small \textbf{Reward explanation:} Correct, it is uniquely green.};
\draw[arrow, bend left] ([xshift=-6, yshift=-4]sp.north) to (1.2, 0.9);
\end{tikzpicture}
}
\vskip-0.4em
\caption{Example 1.} \label{fig:ooo_tasks:perceptual}
\end{subfigure}%
\begin{subfigure}[b]{0.5\textwidth}
\centering
\resizebox{0.95\textwidth}{!}{
\begin{tikzpicture}
\fill[bblue!80!white] (-0.5, -0.75) rectangle (0.5, 0.75);
\node[fill=bred!70!black, circle, minimum height=1cm] at (1.75, 0) (sp) {};
\node[fill=bblue!80!white, circle, minimum height=1cm] at (3.5, 0) {};
\draw[draw=white, pattern=north east lines, pattern color=white] (-1,-1) rectangle (1,1);
\fill[bred!70!black] (4.77, -0.5) rectangle (5.33, 0.5);
\draw[draw=white, pattern=north east lines, pattern color=white] (4.5,-0.6) rectangle (5.5,0.6);

\node[align=left, anchor=west] at (-1.4, 1.8) (exp_lp) {\small \textbf{Properties explanation:} This is a large, striped, blue rectangle.};
\draw[arrow, bend left] (-0.5, 0.75) to (-0.5, 1.65);

\node[align=left, anchor=west, text width=7 cm] at (0, 1.1) (exp_sp) {\small \textbf{Reward explanation:} Incorrect, others are red, \phantom{blahblahblahblahblahbl} solid, small, or circles.};
\draw[arrow, bend left] ([xshift=-6, yshift=-2]sp.north) to (1, 1.1);
\end{tikzpicture}
}
\vskip-0.8em
\caption{Example 2.} \label{fig:ooo_tasks:perceptual2}
\end{subfigure}%
\caption{Conceptual illustrations of two possible odd-one-out tasks, and corresponding possible explanations. This figure depicts odd-one-out tasks with feature dimensions of color, texture, shape, and size, and the two types of explanations we consider. Property explanations identify relevant object features, while reward explanations specify which feature(s) make the choice correct or incorrect.  (\subref{fig:ooo_tasks:perceptual}) The second object is the odd one out, because it is a unique color. (\subref{fig:ooo_tasks:perceptual2}) The first object is the odd one out, because it is uniquely large. Explanations of incorrect choices identify all features.} \label{fig:ooo_tasks}
\end{figure*}

\section{The odd-one-out tasks} \label{sec:tasks}
We first outline a challenging family of fundamentally-relational tasks: finding the odd one out in a set of objects, i.e.\ the one that is somehow unique (Fig.\ \ref{fig:ooo_tasks}). Odd-one-out tasks have been used extensively in cognitive science \citep[e.g.,][]{stephens2008one,crutch2009different}, and proposed in perceptual settings in robotics \citep{sinapov2010odd}. These tasks are challenging, because they involve both relational reasoning (same vs. different) and abstraction (identifying uniqueness requires reasoning over all objects, and all dimensions along which objects may be related). Furthermore, these tasks permit informative explanations of relevant dimensions, properties, and relations.

Investigating these challenging and abstract---yet explainable---relational tasks is particularly interesting, because relational reasoning and abstraction are critical human abilities \citep{gentner2003we,penn2008darwin}, but the capacity of deep learning to acquire these skills is disputed \citep{santoro2017simple,santoro2018measuring,geiger2020relational, ichien2021visual, puebla2021can}. 
However, explanations support human relational learning \citep{gentner2008relational,lupyan2008taking, edwards2019explanation}, suggesting that explanations might similarly help machines acquire these skills.

In Fig.\ \ref{fig:ooo_tasks} we conceptually illustrate some odd-one-out tasks. In Fig.\ \ref{fig:ooo_tasks:perceptual} one object is uniquely green, while the rest are purple. We thus denote color as the \emph{relevant} dimension in this episode. Along the other, irrelevant dimensions---shape, texture, and size---attributes appear in pairs, e.g. there are two pentagons and two triangles. These pairs force the agent to consider \emph{all} the objects. If the agent considered only the first three objects it would be unable to tell whether the first object was the odd one out (uniquely large), the second (uniquely green), or the third (uniquely a triangle or uniquely solid textured). Thus, the agent must consider all objects to identify the correct dimension and the unique feature. This makes the relational reasoning particularly challenging, since the agent must consider many possible relationships. The agent is rewarded for selecting the 
odd-one-out, by picking it up or touching it.

We emphasize that in principle these tasks can be learned from reward alone---language is not necessary for performing them, and we evaluate without language. Nevertheless, we find that in practice language explanations are critical for learning these tasks in our settings. 
We consider two types: reward explanations and property explanations (see Fig.\ \ref{fig:ooo_tasks}). Reward explanations are produced after the agent chooses, and identify the feature(s) that make the choice correct or incorrect. Property explanations are produced before the agent chooses, and explain the identity of the object the agent is facing by specifying its task-relevant properties. Both types satisfy our criterion for explanations: they link the situation and the agent's behavior to the task structure.

\textbf{Environments:}
Odd-one-out tasks can be instantiated in various settings, from games to language or images, and can incorporate various latent structures (e.g. meta-learning). We instantiate these tasks in 2D and 3D RL environments (Fig.\ \ref{fig:basic:envs}). In 2D, the agent has simple directional movement actions, while in 3D it can move, look around, and grasp nearby objects at which it is looking.
In both environments we place an agent in a room containing four objects, which vary along feature dimensions of color, texture, position, and either shape (2D) or size (3D). In each episode, one object will be unique along one dimension. The 3D environment compounds the difficulty of the odd-one-out tasks, because the agent's limited view often forces it to compare objects in memory.  See Appx. \ref{app:methods:environments} for full details.

\section{Method: generating explanations} \label{sec:generating_explanations}

\begin{figure}
\centering
    \resizebox{0.35\textwidth}{!}{
    \begin{tikzpicture}
    \draw[boundingbox, dashed, draw=lightgray] (-2.9, 3.4) rectangle (3.4, 4.1);
    \node[text=lightgray] at (2.65, 3.75) {Encoders};
    \node at (0, 3.15) {Image};
    
    \node[block, minimum width=1.5cm] at (0, 3.75) (venc) {ResNet};
    
    \draw[boundingbox, dashed, draw=lightgray] (-2.9, 4.3) rectangle (3.4, 5.4);
    \node[text=lightgray] at (2.7, 4.5) {Memory};
    
    \node[block, minimum width=1.5cm] at (0, 4.8) (att1) {GTrXL};
    \path[arrow, very thick] (venc.north) to (att1.south);
    \node[block, minimum width=0.5cm, gray] at (-2.4, 4.8) (past2) {\phantom{t}};
    \path[arrow, very thick, bend right, gray] ([xshift=1,yshift=-1]att1.north west) to ([xshift=-1,yshift=-1]past2.north east);
    \node[block, minimum width=0.5cm, gray] at (-1.5, 4.8) (past) {\phantom{t}};
    \path[arrow, very thick, bend right, gray] ([xshift=1,yshift=-1]att1.north west) to ([xshift=-1,yshift=-1]past.north east);
  
    \draw[boundingbox, dashed, draw=lightgray] (-2.9, 5.6) rectangle (3.4, 6.6);
    \node[text=lightgray] at (2.85, 5.8) {Heads};
    
    \node[block, minimum width=0.66cm] at (-2.3, 6.25) (pol) {MLP};
    \node[block, minimum width=0.66cm] at (-1.2, 6.25) (val) {MLP};
    \node[block, minimum width=0.66cm] at (0.35, 6.25) (irec) {ResNet};
    \node[block, minimum width=0.66cm] at (2.4, 6.25) (lrec) {LSTM};
    \path[arrow, very thick] (att1.north) to (irec.south);
    \path[arrow, very thick] (att1.north) to ([xshift=-13]lrec.south);
    \path[arrow, very thick] (att1.north) to ([xshift=10]pol.south);
    \path[arrow, very thick] (att1.north) to ([xshift=4]val.south);

    \draw[boundingbox, dashed, draw=bblue!45!lightgray] (-2.9, 6.7) rectangle (-0.7, 7.5);
    \node[text=bblue!50!lightgray] at (-1.8, 7.25) {\(V\)-trace};
    \node at (-2.3, 6.9) (pol) {\(\pi\)};
    \node at (-1.2, 6.9) (val) {\(V\)};
    
    \node at (0.35, 6.9) (irec) {Image};
    \draw[boundingbox, dashed, draw=bgreen!35!lightgray] (-0.6, 6.7) rectangle (1.3, 7.5);
    \node[text=bgreen!40!lightgray] at (0.35, 7.25) {Reconstruct};

    \node at (2.4, 6.9) (irec) {Explanation};
    \draw[boundingbox, dashed, draw=bred] (1.4, 6.7) rectangle (3.4, 7.5);
    \node[text=bred] at (2.4, 7.25) {Generate};
    \end{tikzpicture}}
\caption{RL agent with auxiliary explanation prediction.} \label{fig:agent_schematic}
\vskip-.5cm
\end{figure}
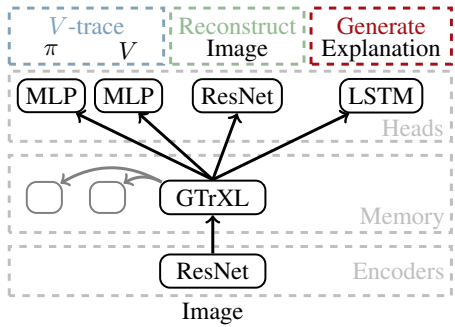
We focus on language explanations provided by the environment during training. We synthetically generate the explanations online, conditional on agent behavior. However,  explanations could be produced by humans, e.g. as annotations of past trajectories \citep[cf.][]{ross2017right}. We train the agent to predict explanations as an auxiliary signal to shape its representations (Fig.\ \ref{fig:agent_schematic}),
as opposed to providing explanations as direct inputs (which is less effective; Appx. \ref{app:analyses:exp_as_input}); our approach thus does not require explanations at test time.
Note that we do not directly supervise behavior through explanations, nor tell the agent how to use them. The agent simply predicts explanations as an auxiliary output.

We train agents using the IMPALA  \citep{espeholt2018impala} framework. Our agent (Fig.\ \ref{fig:agent_schematic}) consists of a visual encoder, a memory, and output heads. The encoder is a CNN or ResNet (task-dependent).
The agent memory is a 4-layer Gated TransformerXL Memory \citep{parisotto2020stabilizing}, which receives the visual encoder output and previous reward as inputs. The output of the memory is input to the heads. The policy and value heads are MLPs, trained with \(V\)-trace. Another head reconstructs the input images to learn better representations (though this is not necessary; Appx.\ \ref{app:analyses:recon_lesion}). Finally, the explanation head is a single-layer LSTM, which generates language explanations. We train the agent to predict these explanations using a summed softmax cross-entropy loss. See Appx. \ref{app:methods:RL} for further agent details.

\section{Experiments}  \label{sec:experiments}

\subsection{Odd-one-out tasks in 2D and 3D RL environments}  \label{sec:experiments:basic}

\begin{figure*}[htb]
\centering
\begin{subfigure}[b]{0.25\textwidth}
\centering
\includegraphics[width=\textwidth]{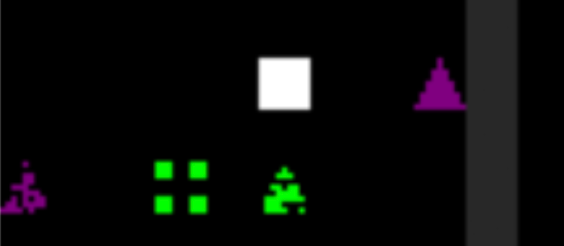}\\[-2pt]
\includegraphics[width=\textwidth]{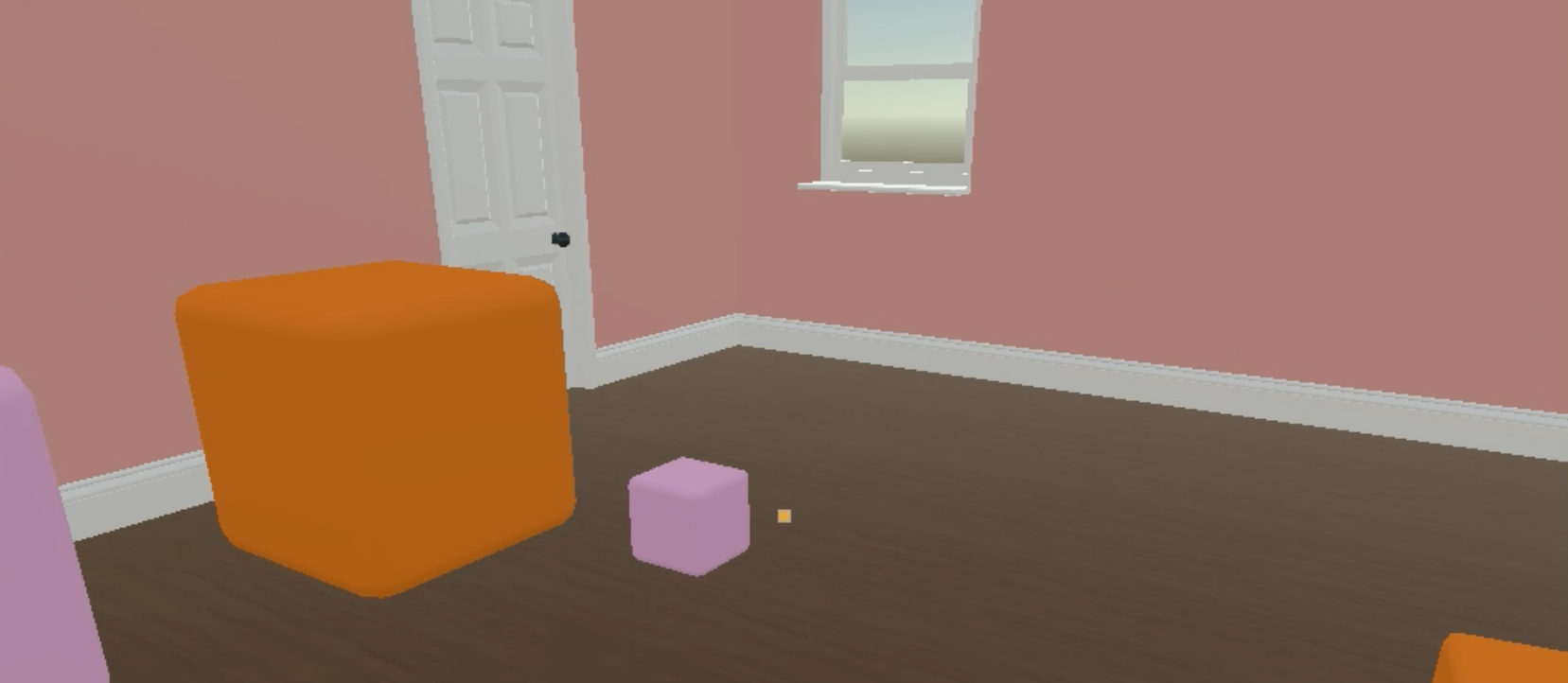}
\vskip0.3em
\caption{2D \& 3D environments.}  \label{fig:basic:envs}
\end{subfigure}%
\begin{subfigure}[b]{0.33\textwidth}
\centering
\includegraphics[width=\textwidth]{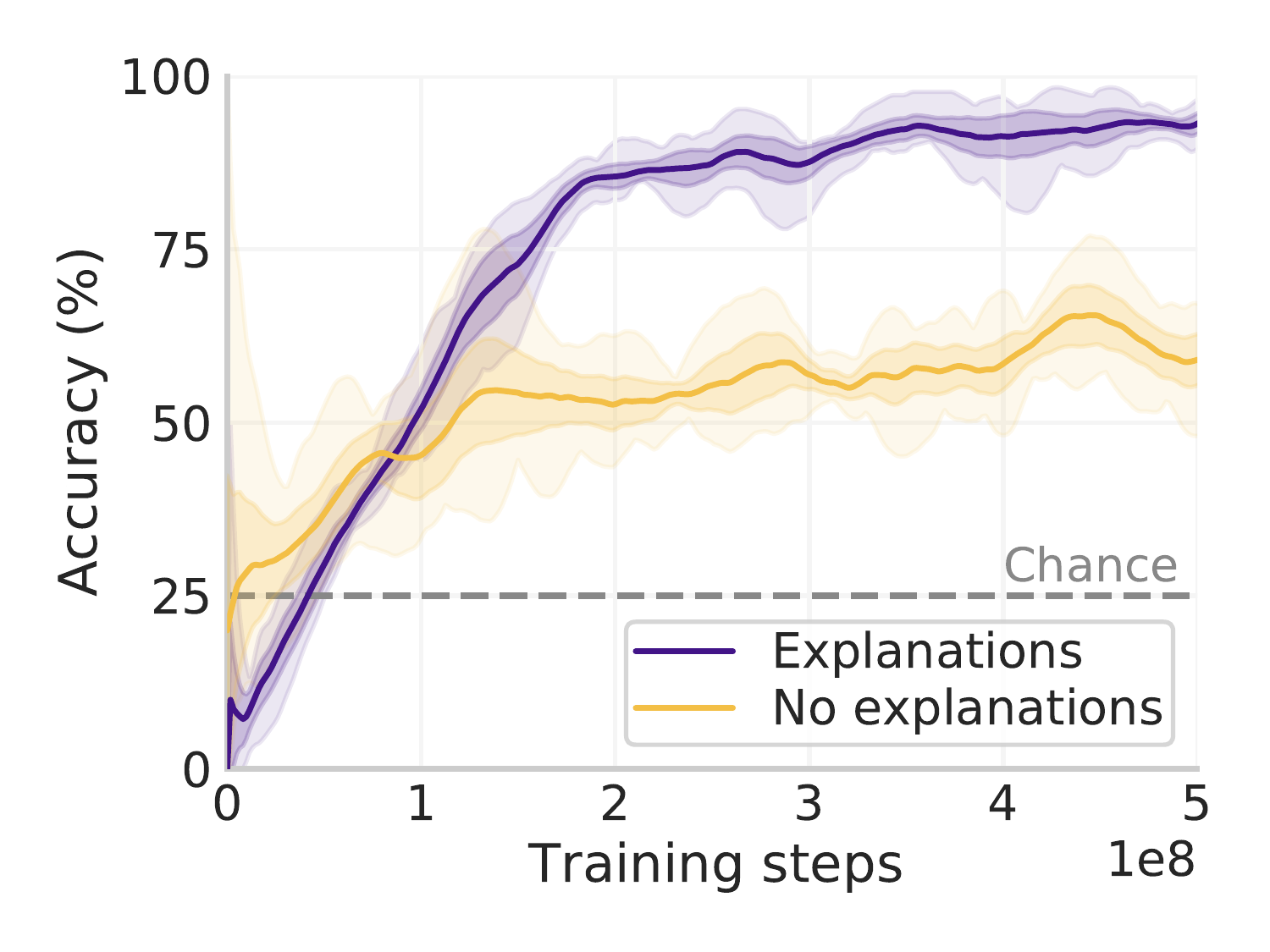}
\vskip-0.5em
\caption{2D results.}  \label{fig:basic:results:pyco}
\end{subfigure}%
\begin{subfigure}[b]{0.33\textwidth}
\centering
\includegraphics[width=\textwidth]{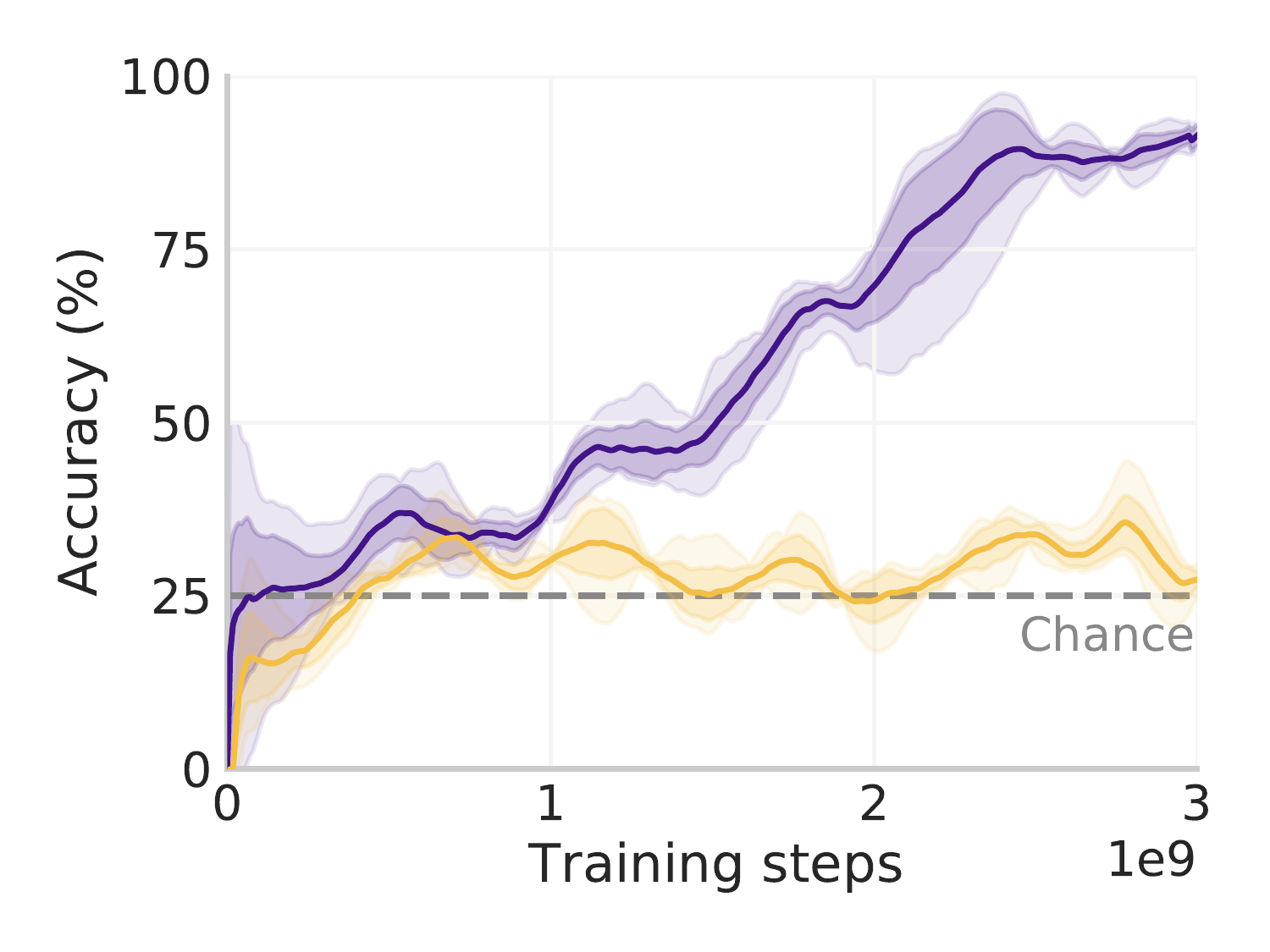}
\vskip-0.5em
\caption{3D results.}  \label{fig:basic:results:playroom}
\end{subfigure}%
\caption{{Explanations help agents learn the perceptual odd-one-out tasks in both RL environments.} (\subref{fig:basic:envs}) Our environments in 2D (top) and 3D (bottom). In 2D, the agent is the white square, while in 3D it has a first-person view. The objects appear in varying positions, colors, textures, and shapes (2D) or sizes (3D). (\subref{fig:basic:results:pyco}) 2D results. Agents trained with explanations achieve high performance; agents trained without explanations do not. (\subref{fig:basic:results:playroom}) 3D results. Only agents trained with explanations exhibit learning substantially above chance. (Training steps denotes actor/environment steps, number of parameter updates is \(\sim10^4\times\) smaller. 5 seeds per condition in 2D, 3 per in 3D, lines=means, dark region=\(\pm\)SD, light region=range.)} \label{fig:basic_envs_results}
\end{figure*}
We first evaluate the benefit of explanations for learning odd-one-out tasks, by comparing agents trained with property and reward explanations to agents trained without. In both 2D and 3D environments, agents trained with explanations learn to solve the tasks over 90\% of the time (Figs. \ref{fig:basic:results:pyco}-\subref{fig:basic:results:playroom}). Agents trained without explanations perform worse; in the easier 2D environment they exhibit partial learning (see \ref{sec:experiments:details}), while in the challenging 3D environment they perform near chance. In 2D all agents were trained with an unsupervised reconstruction loss. However, agents trained without reconstruction but with explanations perform well (Appx.\ \ref{app:analyses:recon_lesion}), while agents trained with reconstruction but without explanations do not. By highlighting abstract task structure, explanations outperform task-agnostic auxiliary objectives, even ones that provide strictly more supervision.

\subsection{Explanations can deconfound} \label{sec:experiments:deconfound}

\begin{figure*}[tb]
\begin{subfigure}[b]{0.34\textwidth}
\centering
\resizebox{\textwidth}{!}{
\begin{tikzpicture}
\node[align=left, anchor=west] at (-1.1, 1.4) {\huge \bf Train (confounded):};
\node[fill=bpurp!80!black,regular polygon, regular polygon sides=3,  minimum height=2cm] at (0, -0.25) {};
\node[fill=bgreen!80!white, regular polygon, regular polygon sides=5, minimum height=1.8cm] at (2, 0) {};
\draw[draw=white, pattern=north east lines, pattern color=white] (1.1,-1) rectangle (2.9,1);
\node[fill=bpurp!80!black,regular polygon, regular polygon sides=3,  minimum height=2cm] at (4,-0.25) {};
\node[fill=bpurp!80!black,regular polygon, regular polygon sides=3,  minimum height=2cm] at (6,-0.25) {};

\node at (10, 0) {\includegraphics[width=4cm]{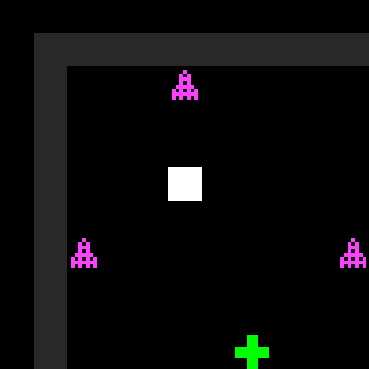}};

\begin{scope}[shift={(0,1)}]
\node[align=left, anchor=west] at (-1.1, -4.2) {\huge \bf Evaluation (deconfounded):};
\fill[bblue!80!white] (-0.5,-6.3) rectangle (0.5,-4.8);
\node[fill=bblue!80!white, circle, minimum height=1.6cm] at (4, -5.56) {};
\draw[draw=white, pattern=north east lines, pattern color=white] (-1,-4.6) rectangle (2.75,-6.7);
\fill[bred!80!black] (1.5,-6.3) rectangle (2.5,-4.8);
\fill[bblue!80!white] (5.5,-6.3) rectangle (6.5,-4.8);
\node at (10, -5.75) {\includegraphics[width=4cm]{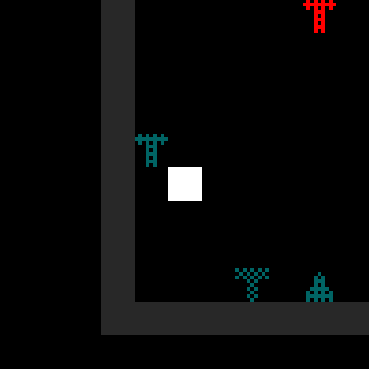}};
\end{scope}
\end{tikzpicture}
}
\vskip0.1em
\caption{Training \& evaluation setup.}  \label{fig:deconfound:task}
\end{subfigure}%
\begin{subfigure}[b]{0.33\textwidth}
\includegraphics[width=\textwidth]{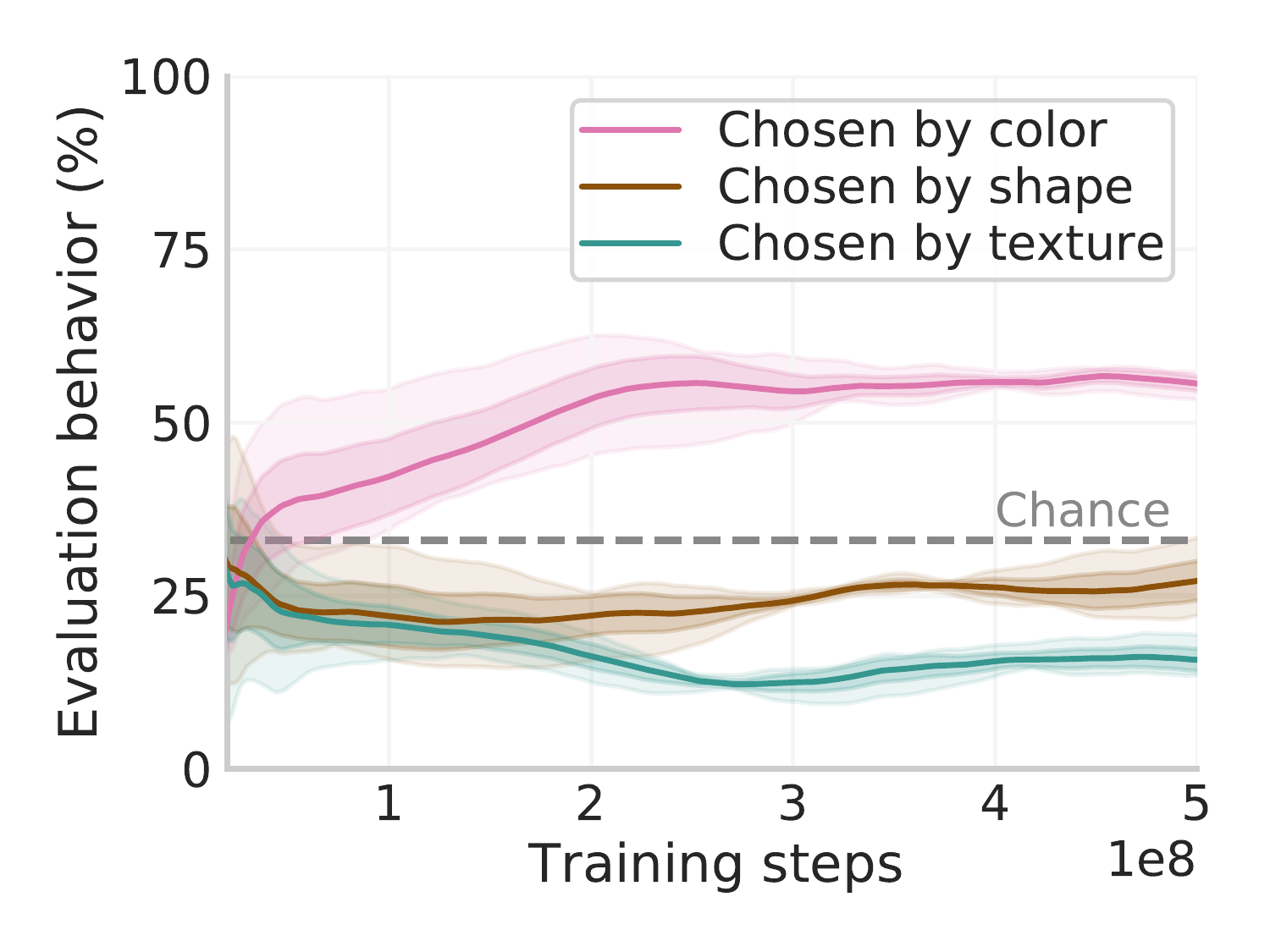}
\vskip-0.5em
\caption{Bias without explanations.}  \label{fig:deconfound:results:noexp}
\end{subfigure}%
\begin{subfigure}[b]{0.33\textwidth}
\includegraphics[width=\textwidth]{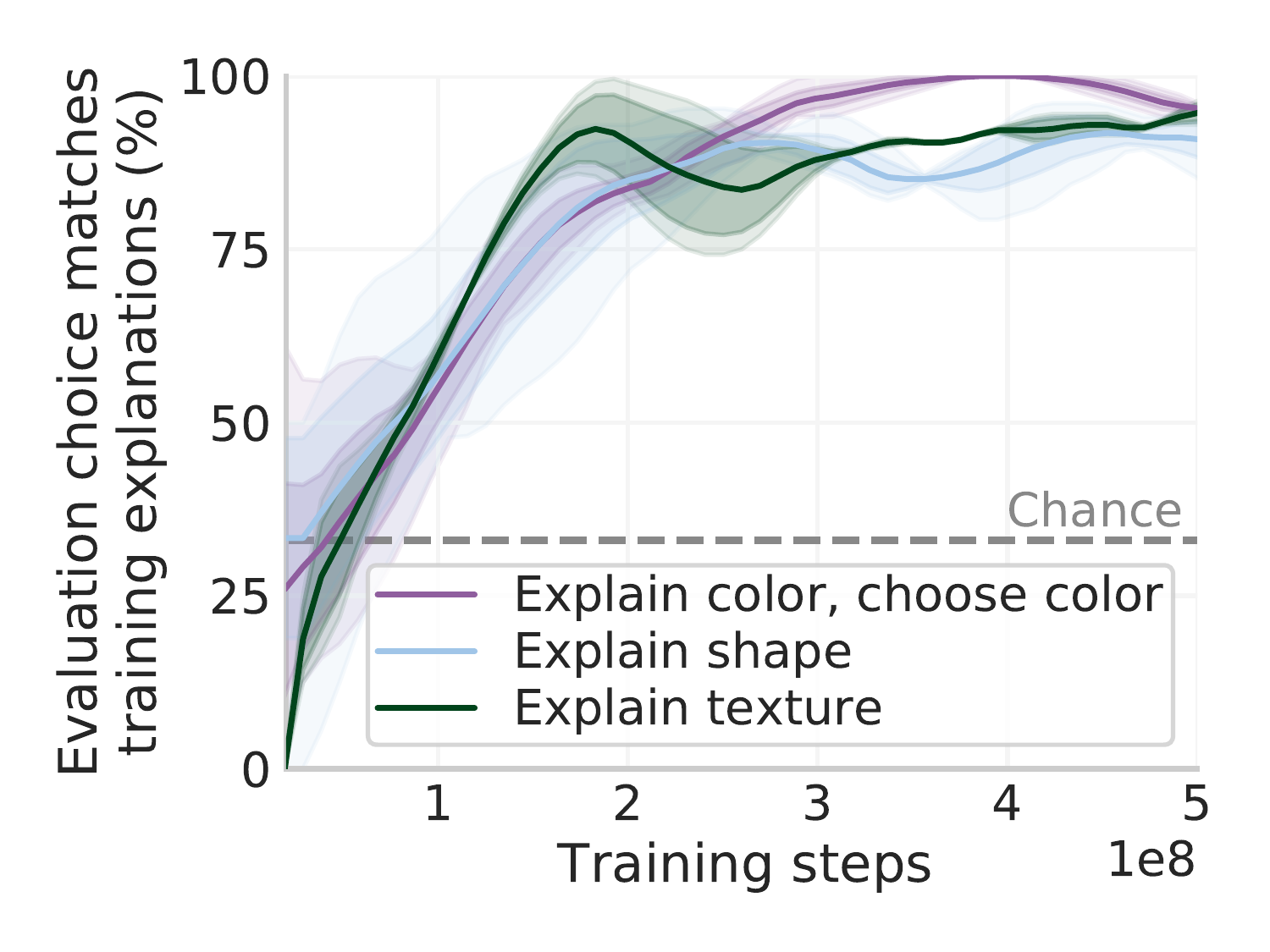}
\vskip-0.5em
\caption{Choices with explanations.}  \label{fig:deconfound:results:explanations}
\end{subfigure}%
\caption{{Explanations can deconfound perfectly correlated features.} (\subref{fig:deconfound:task}) Schematic depictions and environment screenshots from train and test. The agent is trained in confounded settings, where the target object is unique in color, shape, and texture. The agent is tested in deconfounded settings, where one object is unique along each dimension (and an additional distractor object has no unique attributes). (\subref{fig:deconfound:results:noexp}) When trained without explanations, the agent is biased towards using color (the simplest feature) in evaluation. (\subref{fig:deconfound:results:explanations}) However, if the agent is trained with explanations that target any particular feature, the agent prefers that feature in the deconfounded evaluation. (3 seeds per condition, chance is random choice among valid objects.)} \label{fig:deconfound}
\end{figure*}
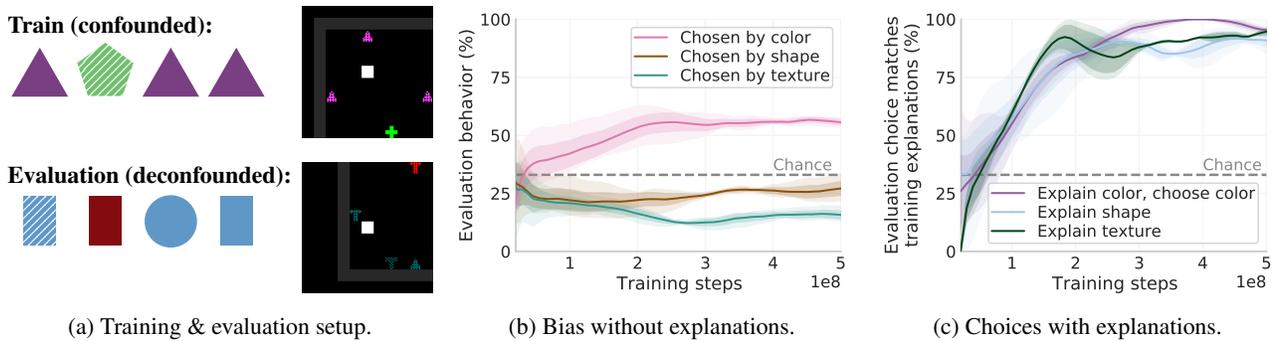

For humans, explanations help identify \emph{which specific aspects} of a situation are generalizable \citep{lombrozo2006functional}. Could explanations also help RL agents to disentangle causally-confounded (perfectly correlated) features, and shape how agents generalize to out-of-distribution tests? We explore this with a different training and testing setup (Fig.\ \ref{fig:deconfound:task}). In training, one object is the odd-one-out along \emph{three} feature dimensions (color, shape, and texture). Thus, any or all of these features could be used to solve the task---the dimensions are perfectly confounded. In test, however, the features are deconfounded: there is a different odd-one-out along each dimension. We explore the effect of explanations that consistently refer to a single feature dimension (without mentioning others) on the agent's behavior in deconfounded evaluation. We train agents in four conditions: no explanations, color-only explanations, shape-only explanations, or texture-only explanations. 
Single-dimension explanations can potentially draw the agent's attention to a particular dimension, and thereby disentangle these features,\footnote{Feature uniqueness is always confounded, but feature values recombine across episodes, allowing disentangling.} even though the explanations do not alter the relationship between these dimensions and the reward signal.

Agents trained without explanations were biased towards using color (the simplest feature) in the deconfounded evaluation (Fig.\ \ref{fig:deconfound:results:noexp}). However, the agents trained with explanations generalized in accordance with the dimension that they were trained to explain \(>85\%\) of the time (Fig.\ \ref{fig:deconfound:results:explanations}), even though there were no direct cues linking the reward to that dimension over the others. In this setting, shaping an agent's internal representations through explanations draws its attention to the desired dimension, and allows \(>85\%\) out-of-distribution generalization along that dimension.

\subsection{Explanations allow agents to learn to experiment} \label{sec:experiments:meta}
\begin{figure*}[htb]

\begin{subfigure}[b]{0.34\textwidth}
\centering
\resizebox{0.95\textwidth}{!}{
\begin{tikzpicture}

\node at (-3, 10.5) (pretrans) {\includegraphics[width=4cm]{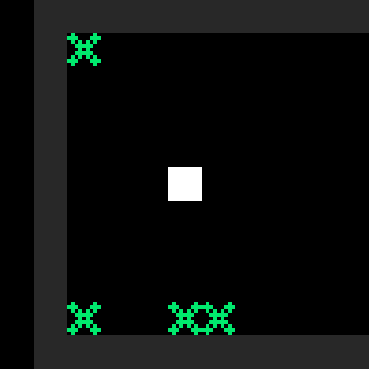}};
\node at (5, 10.5) (posttrans) {\includegraphics[width=4cm]{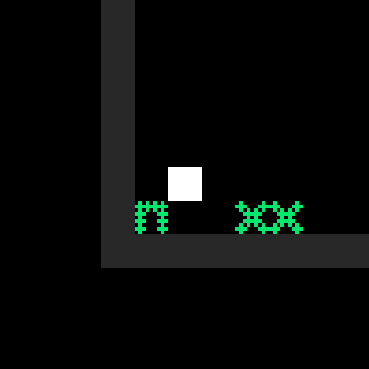}};
\path[arrow, line width=1mm] (pretrans.east) -- node[text width=3.5cm] {\Large Move to object,\\\phantom{blah}\\transform shape} (posttrans.west);
\node at (9, 10.5) (check) {\Large Check};
\path[arrow, line width=1mm] (posttrans.east) -- (check.west);

\begin{scope}[shift={(0,1.5)}]
\node[rotate=90] at (-4.2, 4) (exp) {\Large \bf Experiment};
\node[rotate=90] at (-4.2, -3.5) (test) {\Large \bf Test!};
\path[arrow, line width=1mm] (exp.west) -- (test.east);
\node[rotate=-90] at (9.5, 0.5) (relevant) {\Large \bf Relevant dimension: color};
\path[arrow, line width=0.75mm] (relevant.south) -- (8, 4.25);
\path[arrow, line width=0.75mm] (relevant.south) -- (8, 2.25);
\path[arrow, line width=0.75mm] (relevant.south) -- (8, -0.25);
\path[arrow, line width=0.75mm] (relevant.south) -- (8, -3);

\node[rotate=90] at (-3.5, 4) {\Large Trial 1};
\node[fill=bgreen!80!black,regular polygon, regular polygon sides=5,  minimum height=2cm] at (0, 4) {};
\node[fill=bgreen!80!black,regular polygon, regular polygon sides=5,  minimum height=2cm] at (2, 4) {};
\node[fill=bgreen!80!black,regular polygon, regular polygon sides=5,  minimum height=2cm] at (4, 4) {};
\node[fill=bgreen!80!black,regular polygon, regular polygon sides=5,  minimum height=2cm] at (-2, 4) {};

\path[draw, line width=1mm] (5.46, 3.1) to (6, 4.6);
\node[star, star points=5, minimum size=1 cm, star point ratio=2.25, fill=yellow!80!borange, rotate=-20] at (6, 4.6) {};

\path[arrow, darkgray!90!white, line width=0.75mm, bend left=40] (2.3, 4.9) to node [yshift=3mm] {\Large \bf Transform shape!} (6.7,4.8);
\node[fill=bgreen!80!black,regular polygon, regular polygon sides=3,  minimum height=2cm] at (7, 3.75) {};

\path[draw=bred, line width=1.3mm] (7.3, 5) -- (7.9, 4.4);
\path[draw=bred, line width=1.3mm] (7.3, 4.4) -- (7.9, 5);

\end{scope}

\begin{scope}[shift={(0,1)}]
\node[rotate=90] at (-3.5, 2) {\Large Trial 2};
\node[fill=borange!90,regular polygon, regular polygon sides=7,  minimum height=2cm] at (0, 2) {};
\node[fill=borange!90,regular polygon, regular polygon sides=7,  minimum height=2cm] at (2, 2) {};
\node[fill=borange!90,regular polygon, regular polygon sides=7,  minimum height=2cm] at (4, 2) {};
\node[fill=borange!90,regular polygon, regular polygon sides=7,  minimum height=2cm] at (-2, 2) {};
\draw[draw=white, pattern=north east lines, pattern color=white] (-3, 1) rectangle (5,3);

\path[draw, line width=1mm] (5.26, 1.25) to (5.8, 2.75);
\node[star, star points=5, minimum size=1 cm, star point ratio=2.25, fill=yellow!80!borange, rotate=-20] at (5.8, 2.75) {};

\path[arrow, dash pattern=on2off2, darkgray!90!white, line width=0.75mm, bend left=30] (4.4, 3) to (6.6, 3);
\node[darkgray!90!white] at (2.45, 3.3) {\Large \bf Transform texture!};
\node[fill=borange!90,regular polygon, regular polygon sides=7,  minimum height=2cm] at (7, 2) {};

\path[draw=bred, line width=1.3mm] (7.3, 3.4) -- (7.9, 2.8);
\path[draw=bred, line width=1.3mm] (7.3, 2.8) -- (7.9, 3.4);
\end{scope}

\begin{scope}[shift={(0,0.5)}]
\node[rotate=90] at (-3.5, 0) {\Large Trial 3};
\node[fill=bpurp!70!white,regular polygon, regular polygon sides=3,  minimum height=2cm] at (0, -0.25) {};
\node[fill=bpurp!70!white,regular polygon, regular polygon sides=3,  minimum height=2cm] at (2, -0.25) {};
\node[fill=bpurp!70!white,regular polygon, regular polygon sides=3,  minimum height=2cm] at (4, -0.25) {};
\node[fill=bpurp!70!white,regular polygon, regular polygon sides=3,  minimum height=2cm] at (-2, -0.25) {};

\path[draw, line width=1mm] (5.46, -1.3) to (6, 0.2);
\node[star, star points=5, minimum size=1 cm, star point ratio=2.25, fill=yellow!80!borange, rotate=-20] at (6, 0.2) {};

\path[arrow, dash pattern=on5off2, darkgray!90!white, line width=0.75mm, bend left=20] (2.2, 0.7) to (6.8, 0.5);
\node[darkgray!90!white] at (1, 1.1) {\Large \bf Transform color!};
\node[fill=bgreen!40!bblue,regular polygon, regular polygon sides=3,  minimum height=2cm] at (7, -0.25) {};

\path[fill=bgreen] (7.2, 0.76) -- (7.5,0.3) -- (8.1, 1.3) -- (7.5,0.6) -- cycle;

\end{scope}
\node[rotate=90] at (-3.5, -2) {\Large Trial 4};
\fill[bred!80!black] (-2.75,-2.75) rectangle (-1.25,-1.25);
\fill[bred!80!black] (-0.75,-2.75) rectangle (0.75,-1.25);
\node[fill=bred!80!black, circle, minimum height=1.6cm] at (4, -2.01) {};
\draw[draw=white, pattern=north east lines, pattern color=white] (-1,-2.8) rectangle (2.75,-1.2);
\fill[bblue!80!white] (1.26,-2.75) rectangle (2.75,-1.25);

\path[fill=bgreen] (2.7, -1.24) -- (3,-1.7) -- (3.6, -0.7) -- (3,-1.4) -- cycle;
\node at (6.5, -2) {\huge \bf Choose!};

\end{tikzpicture}
}
\caption{A single intervention trial (top), and multi-trial episode structure (bottom).}  \label{fig:interventions:task}
\end{subfigure}%
\begin{subfigure}[b]{0.33\textwidth}
\centering
\includegraphics[width=\textwidth]{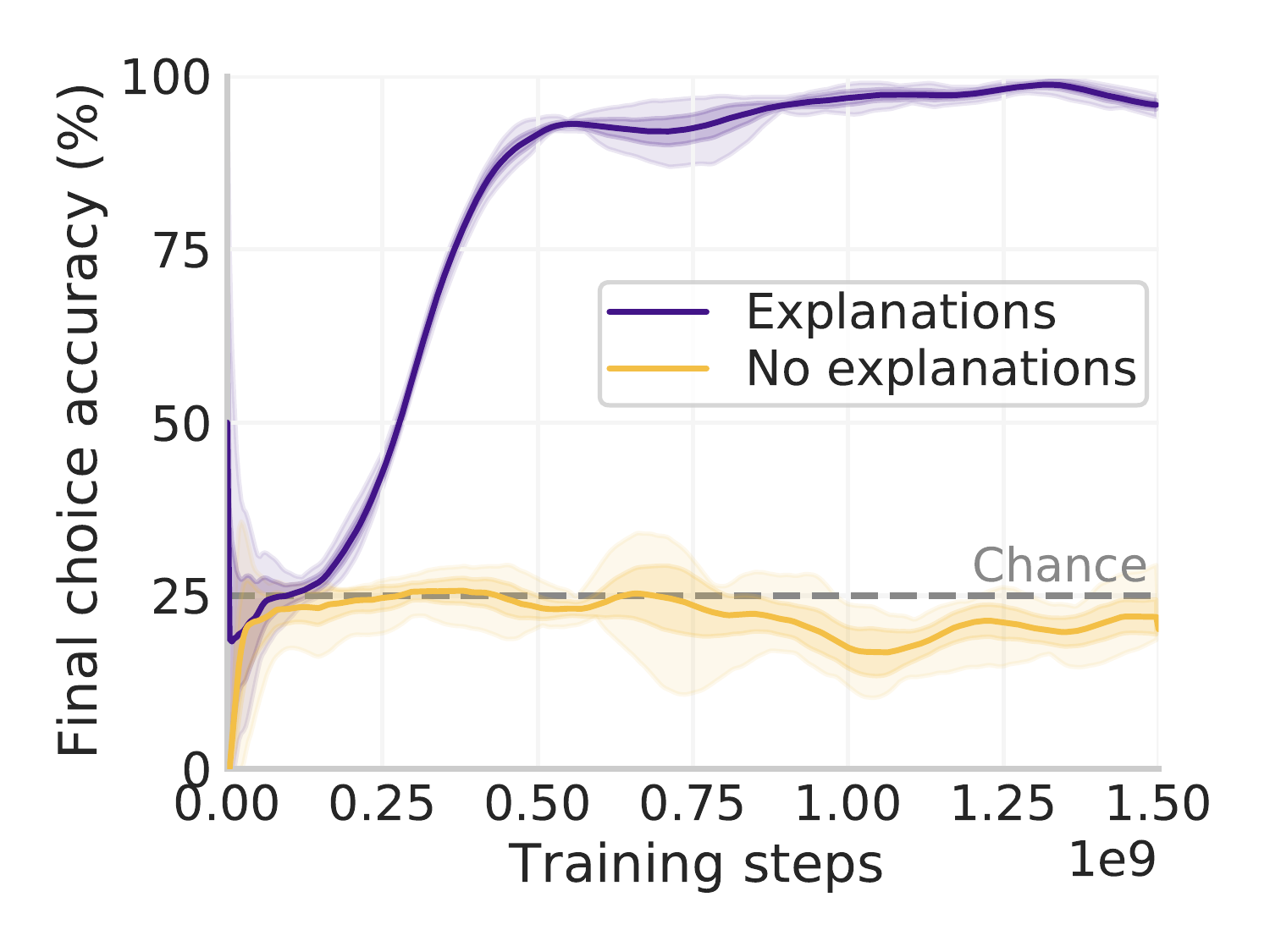}
\resizebox{0.75\textwidth}{!}{
\begin{tikzpicture}
\node[fill=bpurp!70!white,regular polygon, regular polygon sides=3,  minimum height=2cm] at (0, -0.25) {};
\node[fill=bpurp!70!white,regular polygon, regular polygon sides=3,  minimum height=2cm] at (2, -0.25) {};
\node[fill=bpurp!70!white,regular polygon, regular polygon sides=3,  minimum height=2cm] at (4, -0.25) {};
\node[fill=bpurp!70!white,regular polygon, regular polygon sides=3,  minimum height=2cm] at (-2, -0.25) {};

\path[draw, line width=1mm] (5.53, -0.8) to (6, 0.5);
\node[star, star points=5, minimum size=1 cm, star point ratio=2.25, fill=yellow!80!borange, rotate=-20] at (6, 0.5) {};

\path[arrow, dash pattern=on5off2, darkgray!90!white, line width=0.75mm, bend left=40] (4.1, 0.8) to node[yshift=10] {\Large \bf Transform color!} (6.9, 0.8);
\node[fill=bgreen!40!bblue,regular polygon, regular polygon sides=3,  minimum height=2cm] at (7, -0.25) {};

\path[fill=bgreen] (7.2, 0.76) -- (7.5,0.3) -- (8.1, 1.3) -- (7.5,0.6) -- cycle;
\end{tikzpicture}
}
\captionsetup{width=.8\textwidth}
\caption{Easy level structure (bottom) and results (top).}  \label{fig:interventions:results:easy}
\end{subfigure}%
\begin{subfigure}[b]{0.33\textwidth}
\centering
\includegraphics[width=\textwidth]{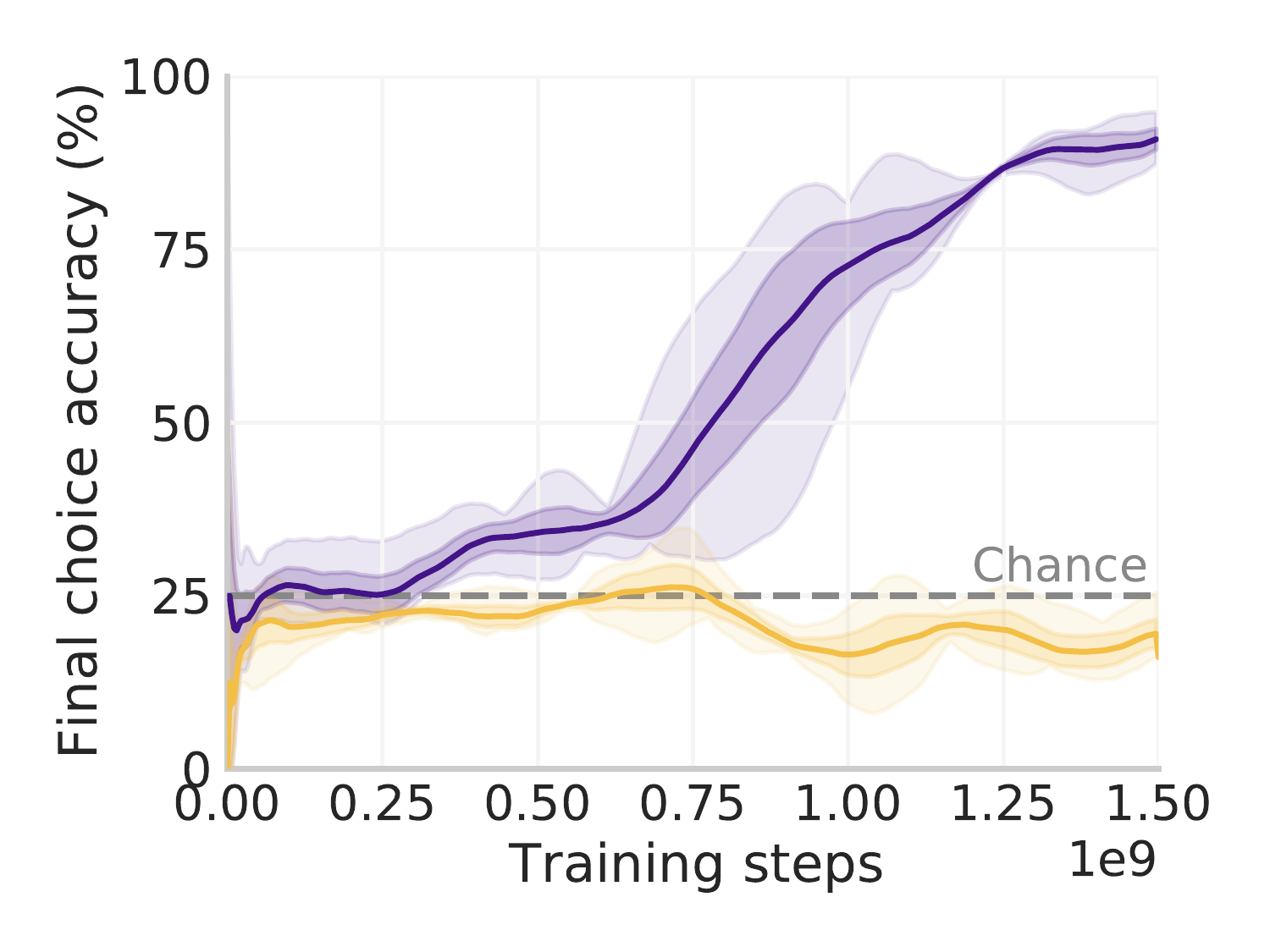}
\resizebox{0.75\textwidth}{!}{
\begin{tikzpicture}
\node[fill=bpurp!70!white,regular polygon, regular polygon sides=3,  minimum height=2cm] at (0, -0.25) {};
\node[fill=bred!70!black,regular polygon, regular polygon sides=5,  minimum height=2cm] at (2, 0) {};
\node[fill=bpurp!70!white,regular polygon, regular polygon sides=3,  minimum height=2cm] at (4, -0.25) {};
\node[fill=bred!70!black,regular polygon, regular polygon sides=5,  minimum height=2cm] at (-2, 0) {};
\draw[draw=white, pattern=north east lines, pattern color=white] (-3, -1) rectangle (1,1);

\path[draw, line width=1mm] (5.53, -0.8) to (6, 0.5);
\node[star, star points=5, minimum size=1 cm, star point ratio=2.25, fill=yellow!80!borange, rotate=-20] at (6, 0.5) {};

\path[arrow, dash pattern=on5off2, darkgray!90!white, line width=0.75mm, bend left=40] (4.1, 0.8) to node[yshift=10] {\Large \bf Transform color!} (6.9, 0.8);
\node[fill=bred!70!black,regular polygon, regular polygon sides=3,  minimum height=2cm] at (7, -0.25) {};
\path[fill=bgreen] (0.2, 0.76) -- (0.5,0.3) -- (1.1, 1.3) -- (0.5,0.6) -- cycle;
\end{tikzpicture}
}
\captionsetup{width=.8\textwidth}
\caption{Hard level structure (bottom) and results (top).}  \label{fig:interventions:results:hard}
\end{subfigure}
\caption{{Explanations allow agents to meta-learn to perform experiments.} (\subref{fig:interventions:task}) Each episode consists of four trials: three where the agent gets to experiment with a magic wand in order to discover which feature dimension is relevant, followed by a final deconfounded trial where it must choose the unique object along that dimension. In this case the relevant dimension is color. In the first trials the agent transforms the shape and texture of the objects, but is not rewarded for picking them up (red X). In the third trial, it transforms the color and is rewarded for picking the object up (green check). The agent can then infer that it should choose the different-colored object in the final trial. (\subref{fig:interventions:results:easy}) In some episodes, the experiments are easy, because all the object attributes are the same, and the agent only needs to transform an object and select that object. Agents trained with explanations learn these tasks, while agents trained without explanations do not. (\subref{fig:interventions:results:hard}) In other episodes, the experiments are harder, because the object attributes are all paired---the agent must transform one object, and then pick up \emph{another} which has been made unique. With explanations, agents learn these difficult levels as well. (4 seeds per condition.)} \label{fig:interventions}
\end{figure*}

Explanations help humans to understand causal structure \citep{lombrozo2006structure,lombrozo2006functional}. The ability of deep learning to infer causality is sometimes questioned \citep[e.g.][]{pearl2019seven}, but while theoretical limitations hold for passive learners, RL agents can intervene and can therefore identify causal structure. Indeed, agents can meta-learn causal reasoning in simple settings \citep{dasgupta2019causal} where causal variables are directly observable and actions allow strong interventions. We investigate whether explanations could help agents learn to identify causal structure in more challenging relational tasks in richer environments. 

We consider a meta-learning setting where agents complete episodes composed of four odd-one-out trials. In each episode, there is only one causally-important dimension in all four trials---reward is determined by uniqueness along only one of the feature dimensions (e.g. color). This ``correct'' dimension changes across episodes, and is not directly observable. Thus agents must learn to \emph{perform experiments} on the first three trials to identify the causally-relevant dimension, in order to select the correct object on a fourth test trial (Fig.\ \ref{fig:interventions:task}). The agent receives 1 reward for completing an early trial correctly, but 10 reward for completing the final trial correctly. Thus, the agent is incentivized to experiment and discover the correct dimension in the early trials, in order to gain a large reward in the final trial.

To enable experiments, in the first three trials of each episode, we give the agent a magic wand that can perform one causal intervention per trial: changing an object's color, shape, or texture. That is, we endow the agent with three additional actions which transform one of those three properties of an adjacent object. The agent is forced to use the wand to create an odd-one-out, because each trial's initial configuration lacks any objects with unique features---along each dimension the features are either all the same, or appear in pairs. When the features are all the same, the experiments are relatively easy (Fig.\ \ref{fig:interventions:results:easy}): the agent must simply transform an object and then select the same object. When the features are paired, however, the experiments are harder (Fig.\ \ref{fig:interventions:results:hard}): the agent must transform one object, which will change to match other objects \emph{and then it must select another object} that was \emph{formerly} paired with this one, but is now unique. The final trial is always a deconfounded test, where a different object is unique along each dimension, and the magic wand is disabled. On all trials, we reward the agent only if it selects an object which is unique along the ``correct'' dimension. Thus, the agent cannot reliably choose correctly unless it has already experimented with the magic wand to infer the correct dimension.

We again compare agents that receive property and reward explanations to agents that do not, but in this case the explanations are augmented to identify the correct dimension (e.g.,\ ``incorrect, the dimension is shape, and other objects are squares''). Again, while in principle these tasks could be learned from rewards alone, we find that agents trained without explanations cannot learn these tasks. However, agents trained with explanations achieve high success at both easy (Fig.\ \ref{fig:interventions:results:easy}) and hard levels (Fig.\ \ref{fig:interventions:results:hard}).
Explanations can help agents learn to perform causal intervention experiments.

\subsection{Exploring the benefits of explanation in more detail} \label{sec:experiments:details}

In order to better understand the benefits of explanations, we explored our results further in a variety of analyses, ablations, and control experiments. We highlight three intriguing results here, and briefly outline the rest.

\textbf{Explanations help agents overcome biases toward easy features (Appx. \ref{app:analyses:lazy}):} In 2D, agents without explanations fixate on positions and colors, and learn to solve the task only when those dimensions happen to be relevant. Shape and texture are generally not learned at all. This explains the moderate performance without explanations. With explanations, by contrast, agents learn to solve the tasks with any feature. 
Similarly, in the confounded features setting color is preferred without explanations, but again with explanations agents can learn to use other features.
\citet{hermann2020shapes} show similar feature-difficulty rankings for CNNs, and that CNNs lazily prefer easier features. Similarly, \citet{geirhos2020shortcut} discuss ``shortcut features'' that networks prefer, despite the fact that those features do not correctly solve the task. Thus, explanations may help an agent to overcome biases towards easy-but-inaccurate solutions, to escape minima or plateaus, and to master the task.

\begin{figure*}[htb]
\centering
\begin{subfigure}[b]{0.33\textwidth}
\centering
\includegraphics[width=\textwidth]{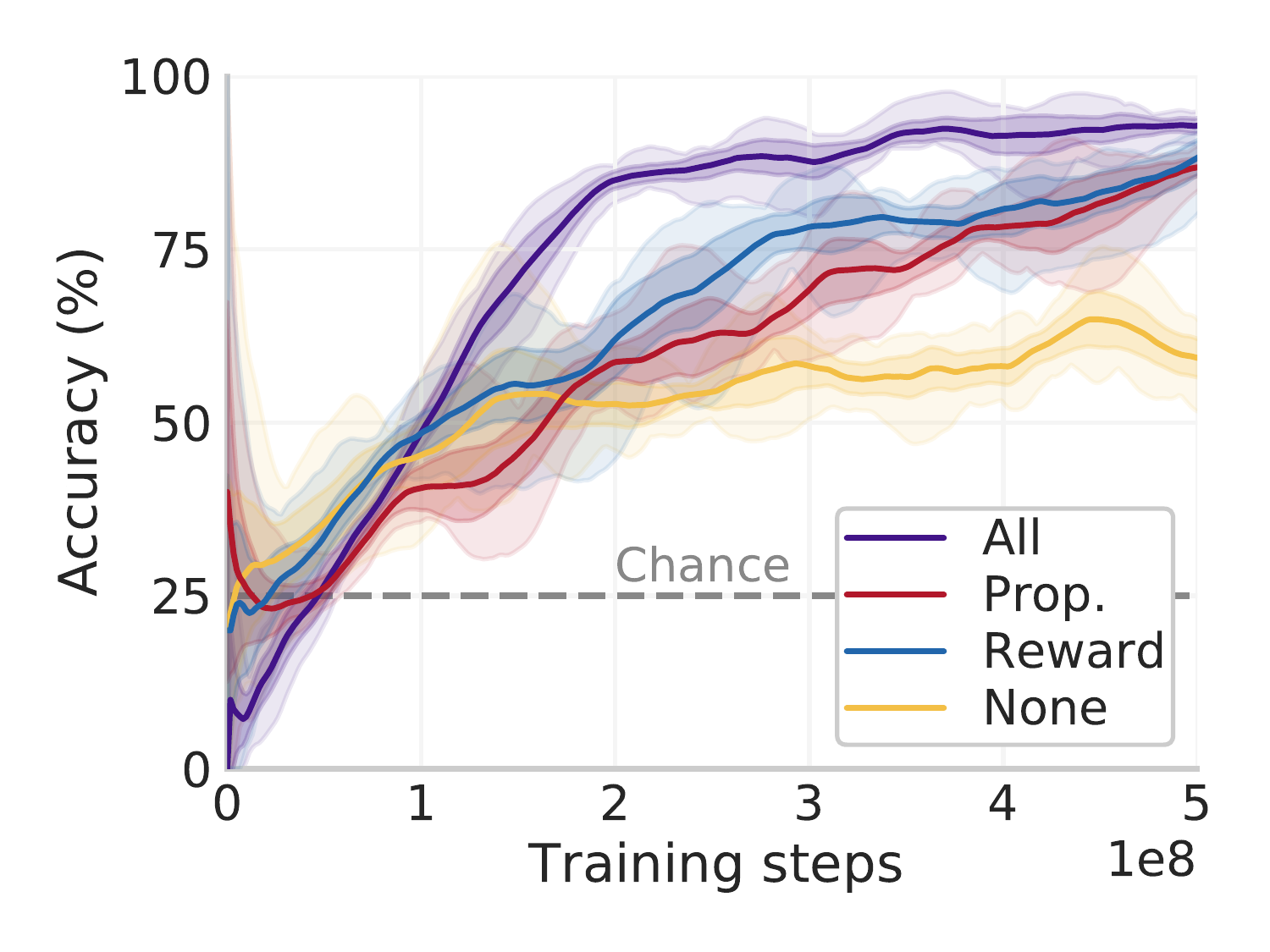}
\vskip-0.25em
\caption{Basic 2D.}  \label{fig:explanation_types:pyco}
\end{subfigure}%
\begin{subfigure}[b]{0.33\textwidth}
\centering
\includegraphics[width=\textwidth]{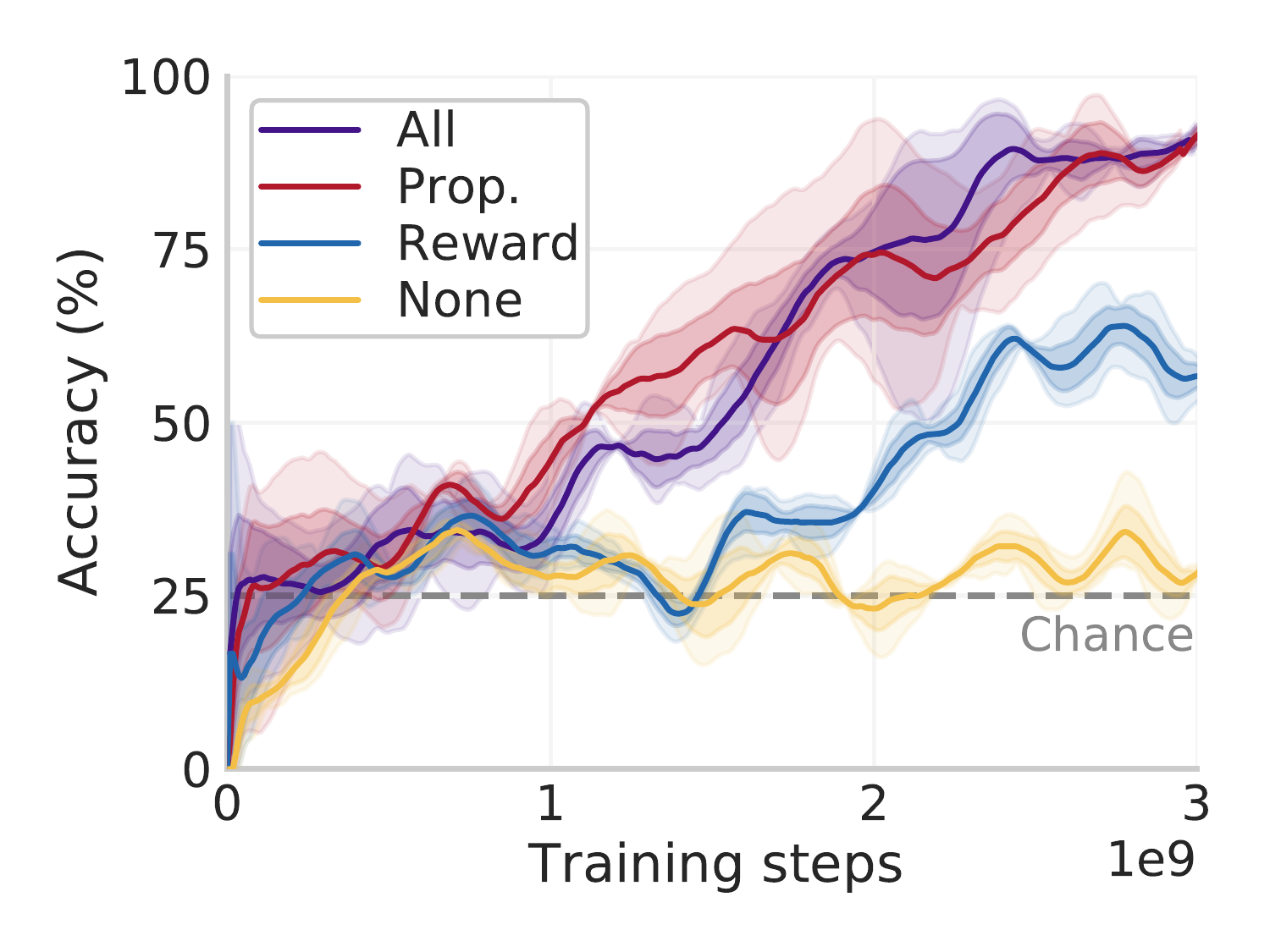}
\vskip-0.25em
\caption{Basic 3D.}  \label{fig:explanation_types:playroom}
\end{subfigure}%
\begin{subfigure}[b]{0.33\textwidth}
\centering
\includegraphics[width=\textwidth]{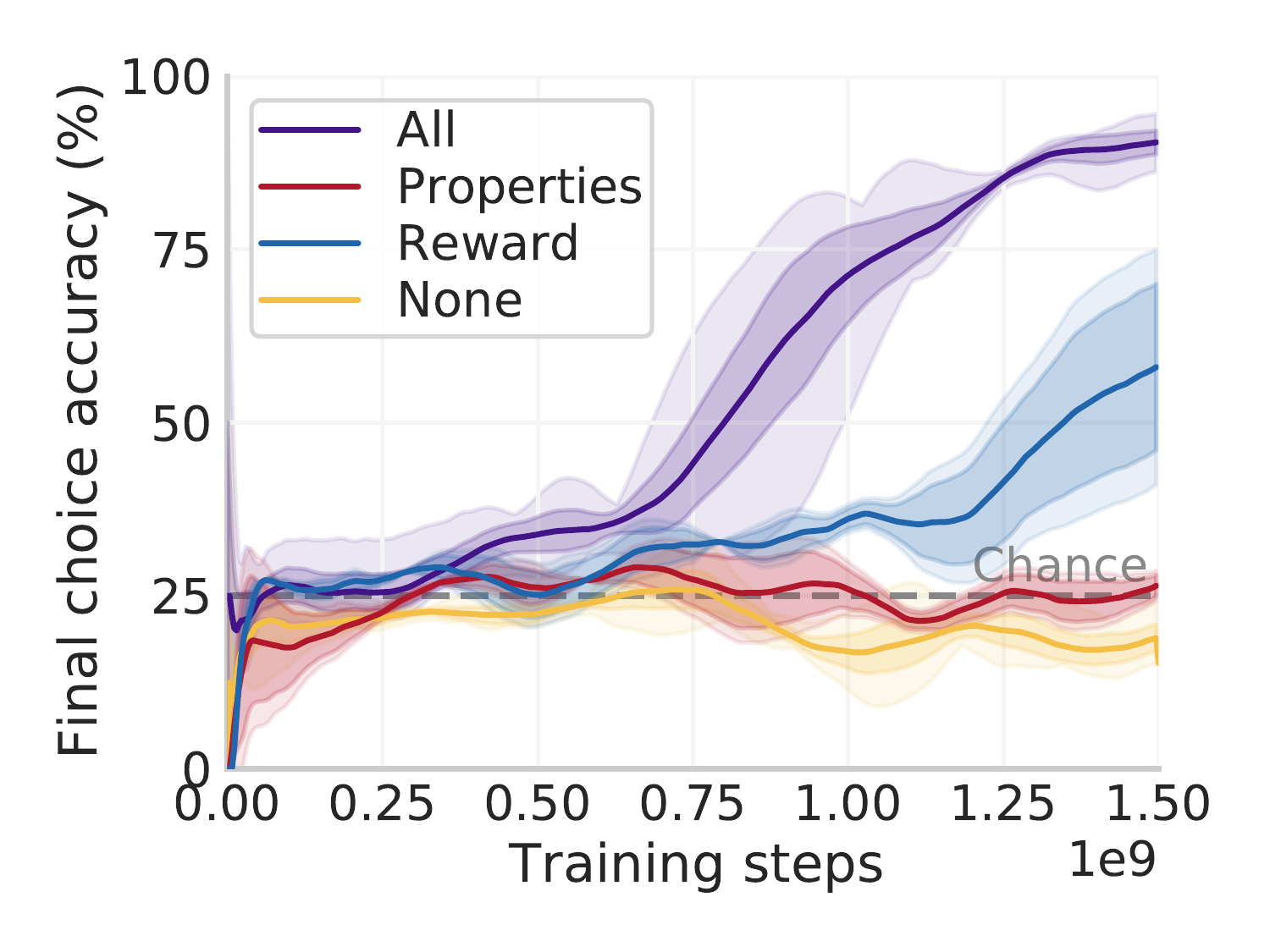}
\vskip-0.25em
\caption{Meta-learning (hard levels).}  \label{fig:explanation_types:meta}
\end{subfigure}%
\caption{Different explanation types offer complementary, separable benefits. We compare agents trained with all explanations or none (as above) to those trained with only property explanations (red), or only reward explanations (blue). (\subref{fig:explanation_types:pyco}) In the basic 2D tasks, either kind of explanations is sufficient for learning, but having both types together is substantially faster. (\subref{fig:explanation_types:playroom}) In the 3D tasks, property explanations result in comparable learning, while reward explanations are not as effective, but still better than none. This is likely due to the memory challenge of these tasks, since it is harder to see all objects at once---property explanations can help the agent discover what to encode to make its choice, while reward explanations cannot. (\subref{fig:explanation_types:meta}) In the learning to experiment setting, by contrast, only reward explanations result in any learning on the hard levels, but both types together is even better. (5 seeds each for All/None conditions, 2 seeds each for properties/reward.)} \label{fig:explanation_types}
\end{figure*}
\textbf{Both explanation types provide complementary benefits; their relative value depends on the environment (Fig.\ \ref{fig:explanation_types}):} In the above experiments we provided agents with both property and reward explanations. Here, we compare to agents trained to generate only a single type of explanations. We found that having both types of explanations is generally better than (or at least as good as) a single type, but the relative benefits of different types depend on the setting. In the 2D environment (Fig. \ref{fig:explanation_types:pyco}) either type of explanations alone results in learning, but both types together result in substantially faster learning. In the 3D setting (Fig. \ref{fig:explanation_types:playroom}), we find that property explanations are relatively more beneficial; perhaps because predicting explanations on encountering the objects helps the agent overcome the memory challenges in the 3D environment by helping it to encode the relevant features in an easily decodable way. By contrast, for the meta-learning tasks (Fig. \ref{fig:explanation_types:meta}), we find that the reward explanations are necessary for any learning. This is likely because the relevance of a transformation experiment to the final reward is much more directly conveyed by the reward explanations than the property ones. However, both types of explanations together are required for complete learning within the training budget we considered. In summary, the relative benefits of the explanation types depend on the demands of the environment, but generally having both types is best.

\textbf{Behaviorally-\ and contextually-relevant explanations are best (Appx. \ref{app:analyses:shuffled}):} Human explanations are \emph{pragmatic} communication: they depend on context, knowledge, and behavior \citep{van1988pragmatic}. We therefore compared to ablation explanations that referred to objects in the room, but independent of behavior (on 10\% of steps we randomly chose an explanation that could occur in the current room, regardless of agent actions), and irrelevant explanations (randomly sampled from those possible in \emph{any} room). We found that behavior-relevant explanations were much more beneficial than behavior-irrelevant ones, and completely irrelevant explanations had no benefit. In particular, in the challenging experimentation tasks, behaviorally relevant explanations were \emph{necessary} for learning. Explanations should engage with the agent rather than passively conveying information. 

\textbf{Other explorations:} 
We briefly describe our other explorations here; see Appx.\ \ref{app:analyses} for full results. We find that language prediction is learned much faster than the RL tasks (Appx. \ref{app:analyses:learning_dynamics}), supporting our suggestion that language makes learning abstractions easier, which can in turn support RL. We found that explanations as input are not helpful (Appx.\ \ref{app:analyses:exp_as_input}), and can even interfere with the benefits of explanations as targets. Prediction is a more powerful signal for learning than receiving an input. We also found that a curriculum of tasks that teaches the agent about object properties---by cueing the agent with a property as input (e.g. ``blue'') and then rewarding the agent if it chooses the matching object---is not as effective as explanations (Appx.\ \ref{app:analyses:curriculum}).
Finally, explanations are more beneficial in complex tasks (Appx.\ \ref{app:analyses:complexity}). Thus, explanations may be especially useful as RL is implied to increasingly complex settings.

\section{Related work}

Language plays a critical role in human learning. Language can identify consistent abstractions or structures in the world, and can shape reasoning processes \citep{edmiston2015makes, lupyan2016centrality,dove2020more}. 
In particular, explanations can enable efficient, generalizable learning, even from a single example \citep{ahn1992schema}. 
Explanations can highlight both causal factors, and relationships between a present situation and broader principles \citep{lombrozo2006structure}. Explanations therefore depend strongly on prior knowledge, and the relationship between explainer, the recipient, and the situation to be explained \citep{van1988pragmatic,cassens2021explanation}. As \citet{wood1976role} say: ``one must recognize the relation between means and ends in order to benefit from ‘knowledge of results.’'' Explanations link a specific situation to more general principles that can be used in the future.

\textbf{Relations:} Relational and analogical reasoning are considered crucial to human intelligence \citep{gentner2003we}, and possibly absent in other animals \citep{penn2008darwin}. The relations \emph{same} and \emph{different} are central to many 
accounts, but their origins are disputed \citep[e.g.]{penn2008darwin, katz2021issues}. But language and culture play a critical role in learning these concepts and skills \citep{gentner2008relational,lupyan2008taking}---``relational concepts are not simply given in the natural world: they are culturally and linguistically shaped'' \citep{gentner2003we}. Thus, explanations may be particularly key, and their absence may help explain neural networks' deficits in relational reasoning \citep{geiger2020relational,puebla2021can,ichien2021visual}, at least without relational inductive biases \citep{santoro2017simple,shanahan2020explicitly}.

\textbf{Causality:} Humans focus on causal structure, even as children \citep{gopnik1999scientist,gopnik2000detecting}, and our causal understanding is closely linked to explanations \citep{lombrozo2017causal}. Human explanations are not just causal, but emphasize important causal factors that are useful for future prediction and intervention \citep{lombrozo2006functional}. Furthermore, \citet{lombrozo2006functional} emphasize that children accept various explanations, while adults selectively endorse causally generalizable ones, suggesting that this focus may be at least partly learned.

\textbf{Self-explanation:} Asking humans to produce explanations for themselves, without providing feedback, can improve generalization \citep[e.g.][]{chi1994eliciting,rittle2006promoting,williams2010role}. Furthermore, \citet{nam2021underlies} find that the ability to produce explanations is strongly related to the ability to learn a generalizable problem-solving strategy involving relational reasoning; and furthermore that education---especially in mathematics---is related to developing these abilities. The skills of explaining and generalizing may be learned together.

\subsection{Related work in AI}

We are certainly not the first to observe that the cognitive literature suggests that explanations might help in AI.
Here we review a variety of prior work on explanation in AI. We also relate to the broader set of approaches for auxiliary supervision that help agents (or models) to learn more effective representations for a task. Explanations are a particularly targeted form of auxiliary supervision that focuses on the causally-relevant, generalizable elements of a situation.

\textbf{Language as representation, or to shape representations?} \citet{andreas2018learning} used language as a latent bottleneck representation in meta-learning, and found benefits. However, \citet{mu-etal-2020-shaping} showed that it was better to \emph{not} bottleneck through language, but merely use descriptions to shape latent representations in supervised classification tasks. We similarly use language as an auxiliary signal to shape latent representations toward task-relevant abstractions. However, we focus on RL, where discovering the right abstractions is more challenging and therefore language can be even more beneficial. RL also allows us to extend to settings like causal intervention, which is not possible in a classification paradigm.

\textbf{Natural Language Processing:} Explanations fit naturally into NLP tasks, and \citet{hase2021can} highlight the many ways that explanations could enter in NLP tasks, e.g.\ as targets, inputs, or as priors. 
They find no improvement from using explanations as targets, but show some positive effects of explanation retrieval during both training and test, including improved performance on relational tasks and better disentangling of causal factors. 

\textbf{Feature explanations as learning tools:} Some prior work has refined models using input attention or gradients as targets for explanatory feedback (e.g. from humans). \citet{ross2017right} show that penalizing gradients to irrelevant features can improve generalization on a variety of image and language tasks.
\citet{lertvittayakumjorn2021explanation} survey works on tuning NLP models using explanatory feedback on features, word-level attention, etc. \citet{schramowski2020making} highlight an intriguing interactive-learning-from-feedback setting where an expert in the loop gives feedback which can be used for similar counter-example- or gradient-based training. \citet{stammer2021right} extend this approach in neurosymbolic models to intervene on symbolically-conveyed semantics rather than purely visual features. In RL, however, applications of feature explanations have been more limited, although \citet{guan2020widening} used human annotations of relevant visual features (and binary feedback) to generate augmentations that varied the task-irrelevant features, and showed benefits over other feature-based explanation techniques or augmentations in video game playing.

\textbf{Language in RL:} Language is used broadly in RL, whether as instructions \citep[e.g.][]{hermann2017grounded,kaplan2017beating}, to target exploration \citep{goyal2019using,watkins2021teachable,mu2022improving,tam2022semantic}, or as an abstraction to structure hierarchical policies \citep{jiang2019language}. \citet{luketina2019survey} review many recent uses of language in RL, and argue for further research. However, they do not even mention explanations. \citet{tulli2020learning} consider natural language explanations of actions in RL. However, they only evaluate a simple, symbolic MDP, and observe no benefits, perhaps because their explanations do not relate to the abstract task structure.

\textbf{Auxiliary tasks:}
Predicting explanations is part of the general paradigm of shaping agent representations with auxiliary signals \citep[e.g.][]{jaderberg2016reinforcement}. However, explanations are fundamentally different from unsupervised losses---unsupervised objectives are task-independent by definition, while explanations selectively emphasize the causally relevant features of a situation, and the relationship to general task principles \citep{lombrozo2006functional, lombrozo2006structure}. 
Some \emph{supervised} auxiliary objectives are more similar to explanations; the boundaries of explanation are blurry. In the Alchemy environment \citep{wang2021alchemy}, which involves learning latent causal structure, predicting task-relevant features 
improves performance. Similarly, \citet{santoro2018measuring} show that predicting a ``meta-target''---an abstract label encoding some task structure---improves learning of a relational reasoning task. More broadly, supervising task inference can improve meta-learning \citep{rakelly19efficient,humplik2019meta}. Since these predictions directly relate to task structure, they are closer to explanations than unsupervised task-agnostic predictions. However, they do not necessarily actively link the details of the present situation to the principles of the task, as human explanations do.

\section{Discussion}

We explored relational and causal tasks that are challenging for RL agents to learn from reward alone. 
In all cases, learning to generate language descriptions and explanations significantly improved performance. Even though our agents lacked prior knowledge of language, they were able to rapidly learn to predict language explanations, and this prediction helped them to discover the reasoning processes necessary for the task. Explanations help agents learn the challenging, essential abilities of relational and causal reasoning.

We particularly emphasize the causal benefits of explanations, because causal understanding is essential to effective generalization. Indeed, explanations shaped how our agents to generalize out-of-distribution \citep[cf.][]{ross2017right} from ambiguous, causally-confounded data; furthermore, explanations enabled agents to learn to perform their own experimental interventions to identify causal structure. Without access to a pre-specified, discrete causal diagram  \citep[e.g.][]{pearl2019seven}, our agents were able to ground explanations in pixel-level inputs to learn and generalize causal structure. Humans use explanations to highlight the causally-relevant factors of a task \citep{lombrozo2006functional}, and our results show that explanations can play that same role for agents. 

This focus on task-specific structure allows explanations to outperform task-agnostic auxiliary objectives.
Indeed, we found that explanations helped agents to move beyond a fixation on easy shortcuts that do not fully solve the task, but that models nevertheless prefer \citep[cf.][]{hermann2020shapes,geirhos2020shortcut}.
Explanations offer a promising route to training RL agents that learn and generalize better.

\textbf{Criteria for explanations:} Explanations should satisfy certain criteria for maximal benefits. Explanations must relate between the context, the agent's behavior, and the abstract task structure---explanations that ignore behavior are less useful, and those that ignore context are useless. Receiving explanations as input was not useful in our experiments, likely because it is easier for the agent to ignore inputs than auxiliary targets. Furthermore, explanations outperform unsupervised auxiliary reconstruction. Thus, simply training agents with more information (as with unsupervised objectives) is often not sufficient; explanations must provide relevant and specific learning targets to be most beneficial. 

Nevertheless, we acknowledge that the boundaries of explanation are vague \citep[cf.][]{van1988pragmatic, woodward2005making, sep-metaphysical-explanation}. For example, descriptions cannot name every property, so they tend to pragmatically focus on causally-relevant ones, and thus highlight similar features to explanations.
This fact is why we refer to task-relevant property descriptions as explanations. Furthermore, we use ``explanation'' to refer to cues to relationships between specific situations, behaviors and abstract principles, which may overlap with other forms of auxiliary supervision. While we focused on language explanations, non-language predictions that highlight abstract task features could likely serve the same purpose. Explanations can also vary in abstraction \citep[cf.][]{fyfe2014concreteness,watkins2021teachable}. The boundaries of explanation should be explored further in future work.

\textbf{Limitations and future directions:} We performed our experiments on variations of odd-one-out tasks in RL, which may seem to limit the breadth of our conclusions. However, our different experiments cover many central challenges, including relational and causal structure, confounding, and meta-learning, as well as 2D and 3D environments. Thus we expect that our results will generalize to other challenging settings in future work; even beyond RL.

In broader settings, the ground truth for explanations may not be automatically accessible. In some cases, knowledge about abstractions such as property descriptions may be accessible through large pretrained models---indeed, in subsequent work we have shown that the abstractions from large language-supervised image captioning systems can improve RL exploration \citep{tam2022semantic}---which may provide a scalable route to language supervision.
However, these models may not have access to causal or task-specific structure, or may not work well in domains like Atari which are outside their training distribution. In such cases it may be necessary to rely on alternative approaches, particularly human annotation. Nevertheless, as noted above, explanation is an especially natural and rich form of feedback for humans to provide, so collecting such data may be worthwhile.

Although we drew inspiration from human uses of explanation, our agents do not learn from explanations in the same way as humans. Humans can use our prior knowledge of language to learn from a single explanation in context. By contrast, our agents needed to learn about language \emph{simultaneously} with learning about the tasks, through many repetitions of similar explanations. Future work should explore whether agents that are trained with language and explanations across a diverse array of tasks can meta-learn how to learn from explanations in a more human-like way.

We also do not want to imply that explanations are \emph{necessary} for learning. Most of our tasks could potentially be learned with sufficient data alone, especially if combined with more complicated techniques, for example data augmentation \citep{raileanu2020automatic,guan2020widening}, or auxiliary generative model learning \citep{gregor2019shaping}. 
Furthermore, many promising domains for deep learning---such as protein structure prediction \citep{jumper2021highly}---are precisely those areas that humans do not understand well, and so are challenging domains for humans to explain. Indeed, some domains might be irreducibly complex; in these domains forcing a system to strictly follow simple explanations could be detrimental. Our approach does not force the agent to use explanations directly, and therefore might be less harmful in such cases than stronger constraints like requiring symbolic representations \citep[e.g.,][]{garcez2020neurosymbolic}. 

In other domains there may exist simple explanations that humans have not yet discovered. This observation motivates a future research direction: learning to explain over diverse task distributions, leveraging human explanations in domains we do understand. 
A curriculum focused on producing explanations could
potentially yield substantial benefits. 
Humans generalize better after explaining, even without feedback \citep{chi1994eliciting,rittle2006promoting}, and this ability may be learned through education \citep[cf.][]{nam2021underlies}.  An agent that similarly learns to produce
explanations might similarly learn to generalize better \emph{even in some domains for which we lack ground truth explanations}, and its explanations might help humans interpret its behavior, and the domains in which it performs.

\textbf{Conclusions:} We considered a challenging set of relational and causal tasks, and showed that learning to predict language descriptions and explanations helps RL agents to learn and generalize these tasks across various settings and paradigms.
Explanations can help agents move beyond biases favoring easy features, determine how agents generalize out-of-distribution from ambiguous experiences, and allow agents to meta-learn to perform experiments to identify causal structure. Because these abilities are challenging for current agents, generating explanations as an auxiliary learning signal---rather than purely for post-hoc interpretation---may be a fruitful direction for further research.

\section*{Acknowledgments}
We thank Antonia Creswell and MH Tessler for comments on the manuscript, as well as several anonymous reviewers.

\bibliography{main}
\bibliographystyle{icml2022}

\newpage
\appendix
\onecolumn

\newpage
In Appendix \ref{app:analyses} we show additional experiments and analyses. In Appendix \ref{app:quantitative} we report numerical values for our main experimental comparisons. In Appendix \ref{app:methods} we report details of the environments, agents, and training.

\section{Ablation experiments \& further analyses} \label{app:analyses} 

In this section, we perform a variety of control, ablation, and auxiliary experiments that identify which attributes of explanations are useful in different settings. We perform most of these experiments in the 2D RL setting because of the efficiency of running and training agents in this environment.

\subsection{Agents trained without explanations fixate on the easiest feature dimensions} \label{app:analyses:lazy}

\begin{figure}[H]
\centering
\begin{subfigure}[b]{0.25\textwidth}
\centering
\includegraphics[width=\textwidth]{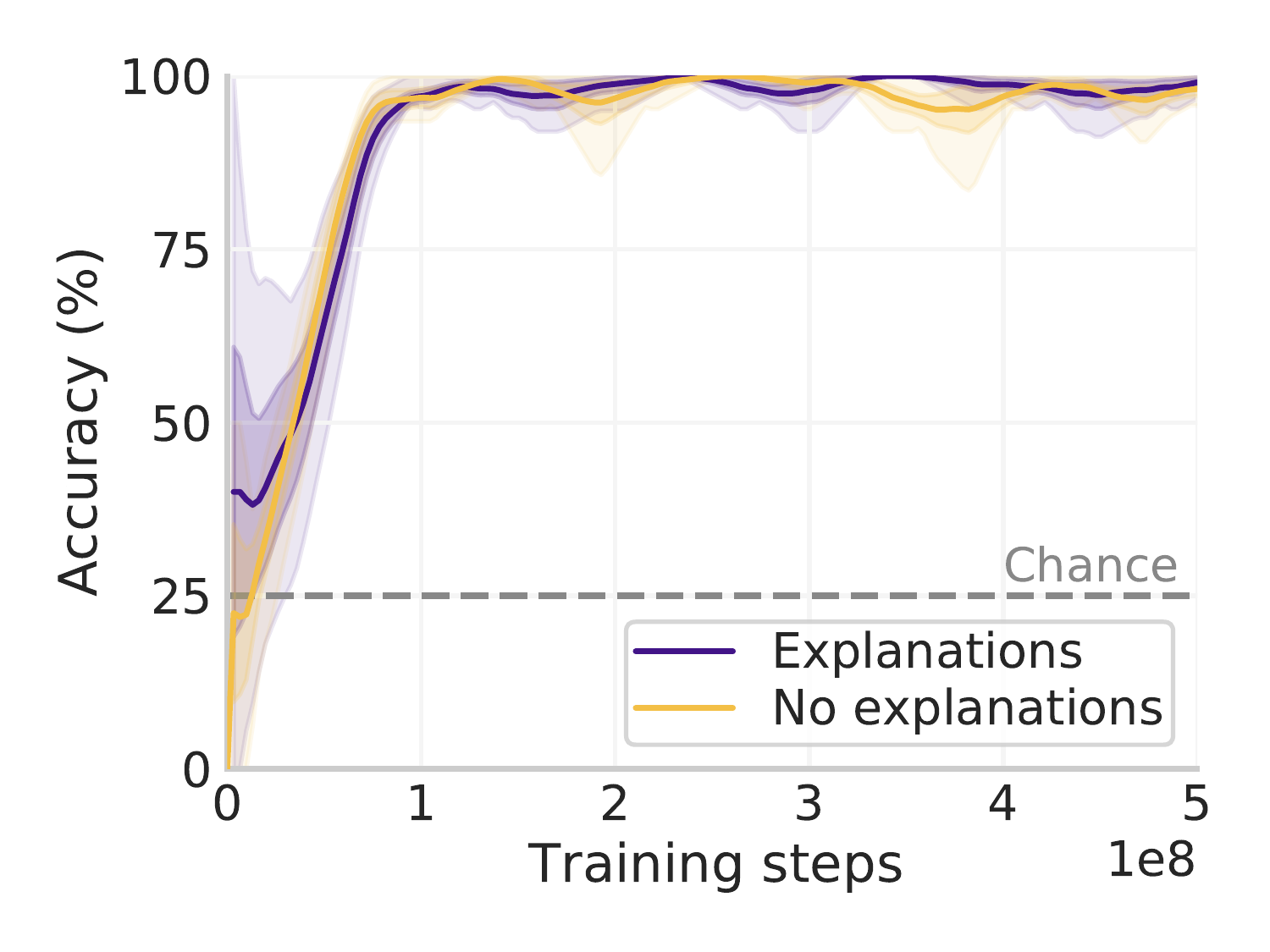}
\caption{Position.}  \label{fig:basic_by_dimension:position}
\end{subfigure}%
\begin{subfigure}[b]{0.25\textwidth}
\centering
\includegraphics[width=\textwidth]{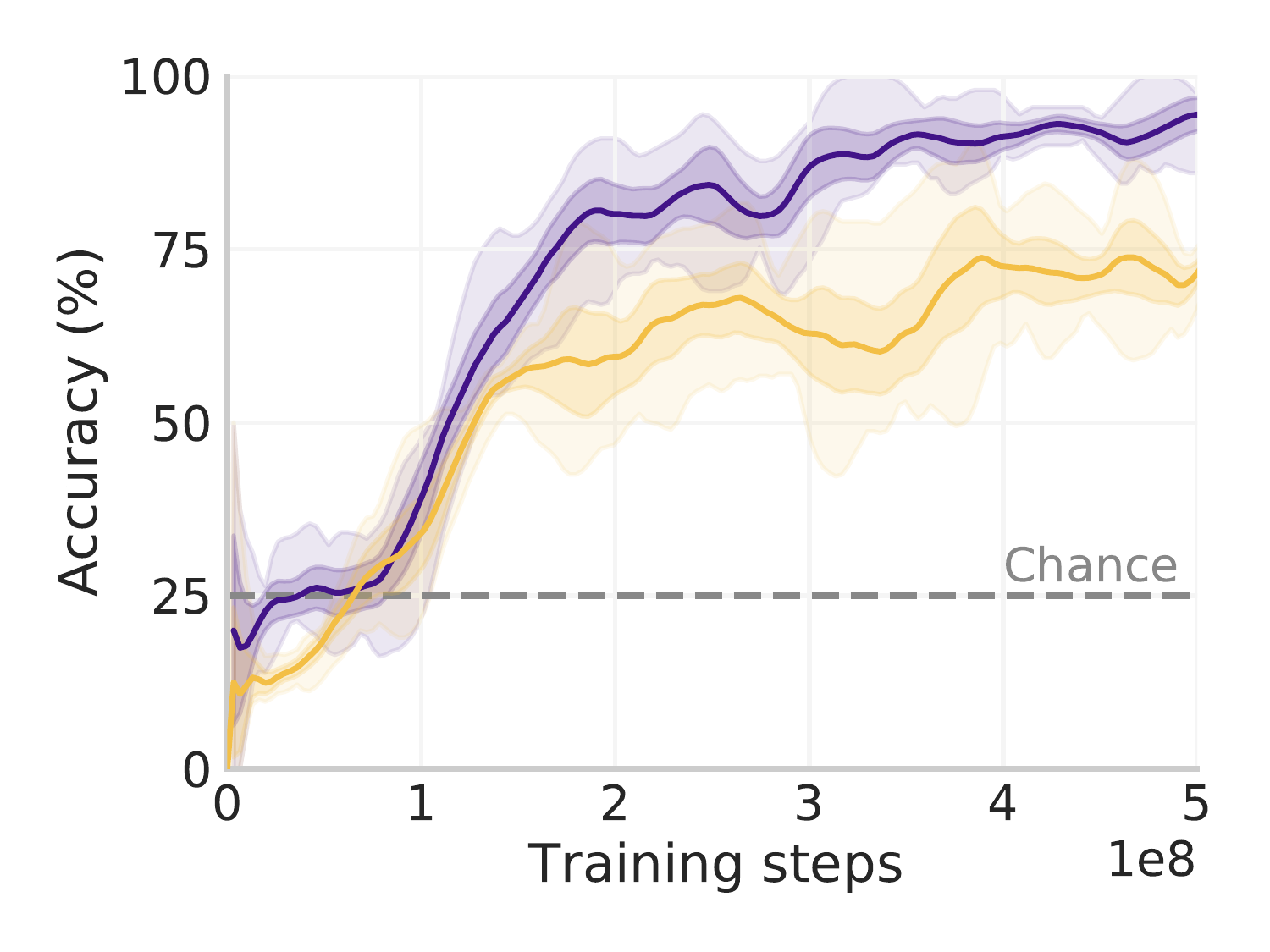}
\caption{Color.}  \label{fig:basic_by_dimension:color}
\end{subfigure}%
\begin{subfigure}[b]{0.25\textwidth}
\centering
\includegraphics[width=\textwidth]{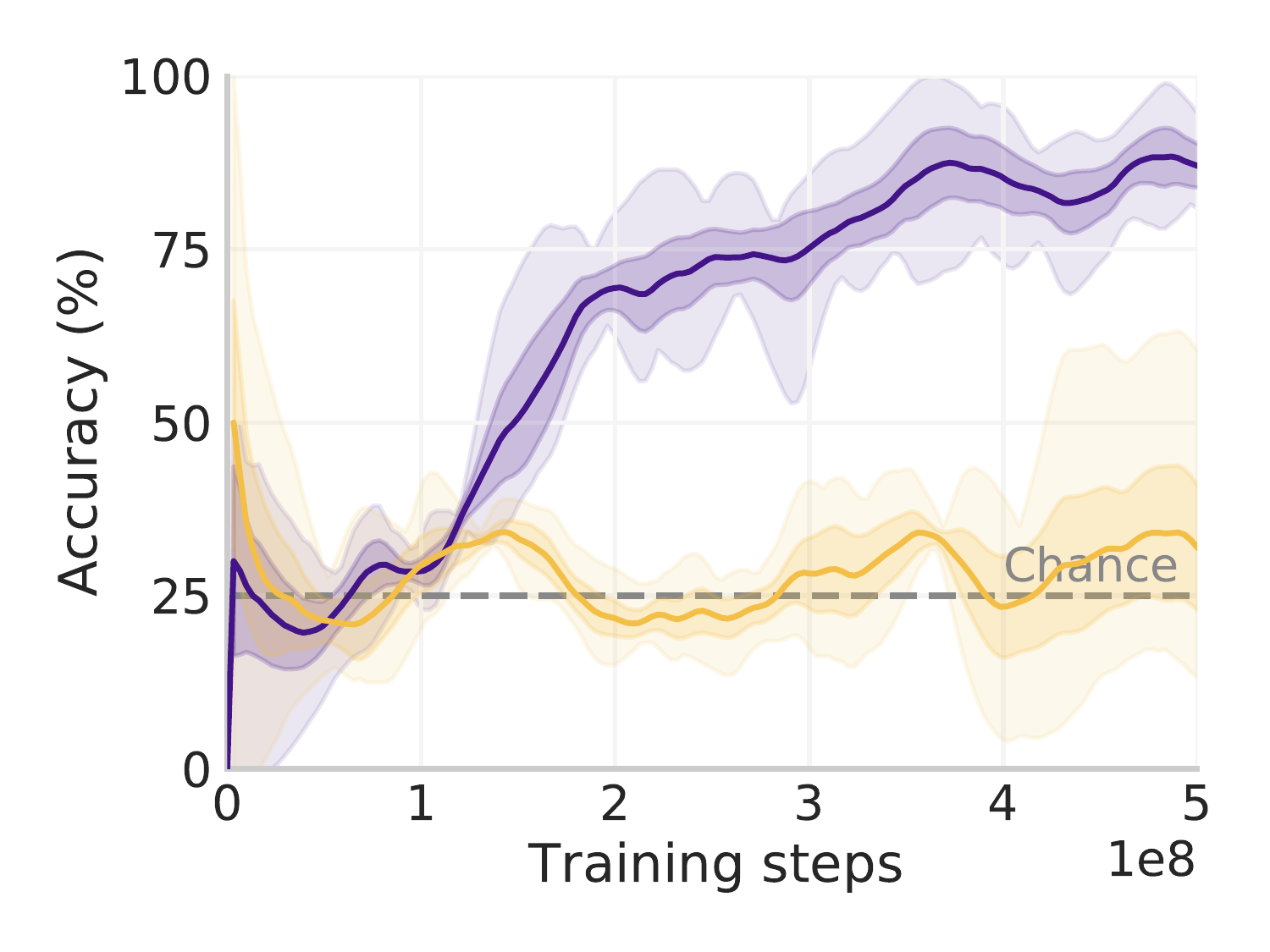}
\caption{Shape.}  \label{fig:basic_by_dimension:shape}
\end{subfigure}%
\begin{subfigure}[b]{0.25\textwidth}
\centering
\includegraphics[width=\textwidth]{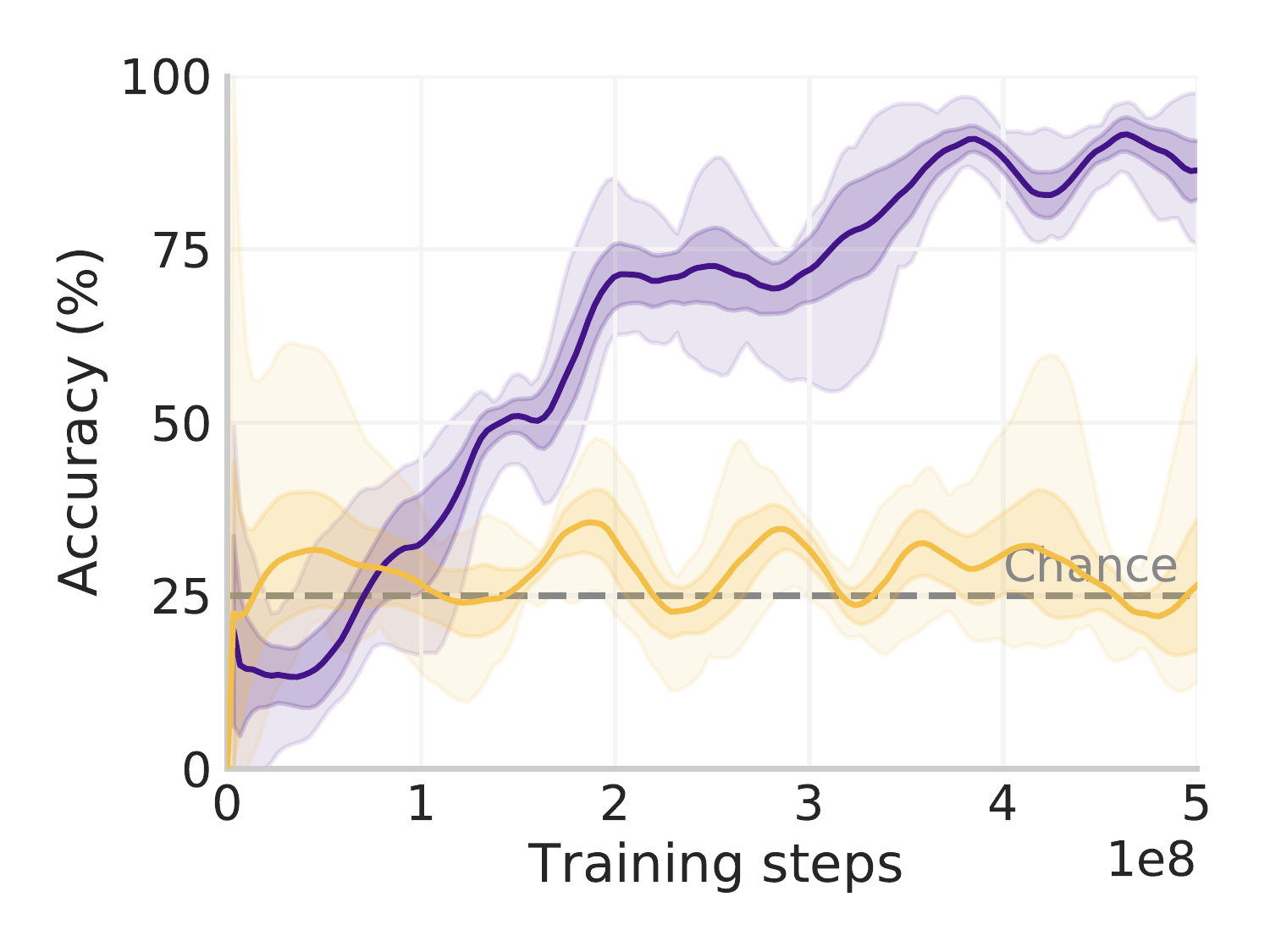}
\caption{Texture.}  \label{fig:basic_by_dimension:texture}
\end{subfigure}%
\caption{In the 2D setting, agents trained with explanations learn all dimensions, but agents trained without explanations learn to fully solve the tasks only if the relevant dimension is position (the easiest dimension), and only partly learn to solve the tasks with color (the next easiest dimension). (5 seeds per condition.)} \label{fig:basic_by_dimension}
\end{figure}

In the basic 2D odd-one-out tasks, the agent achieves off-chance performance without explanations (while in more complicated settings such as the causal interventions, it cannot learn at all without explanations). In Fig.\ \ref{fig:basic_by_dimension}, we show that what the agent is doing is latching on to the feature dimension(s) that are most salient and \emph{easiest} \citep{hermann2020shapes}, and only correctly solving episodes involving these features. Specifically, position is the most salient feature and is learned rapidly even without explanations, followed by color which is partially learned without explanations. However, shape and texture are much more difficult and are not learned well without explanations. These results concord with the features that \citet{hermann2020shapes} found were easiest for CNNs and ResNets to learn, suggesting that explanations may help overcome the preference of agents (or other networks) to be ``lazy'' and prefer ``shortcut features'' \citep{geirhos2020shortcut}. 

\subsection{Explanations are most useful if they engage with the agent's behavior; shuffled explanations are useless} \label{app:analyses:shuffled}

\begin{figure}[htb]
\centering
\begin{subfigure}[b]{0.33\textwidth}
\centering
\includegraphics[width=\textwidth]{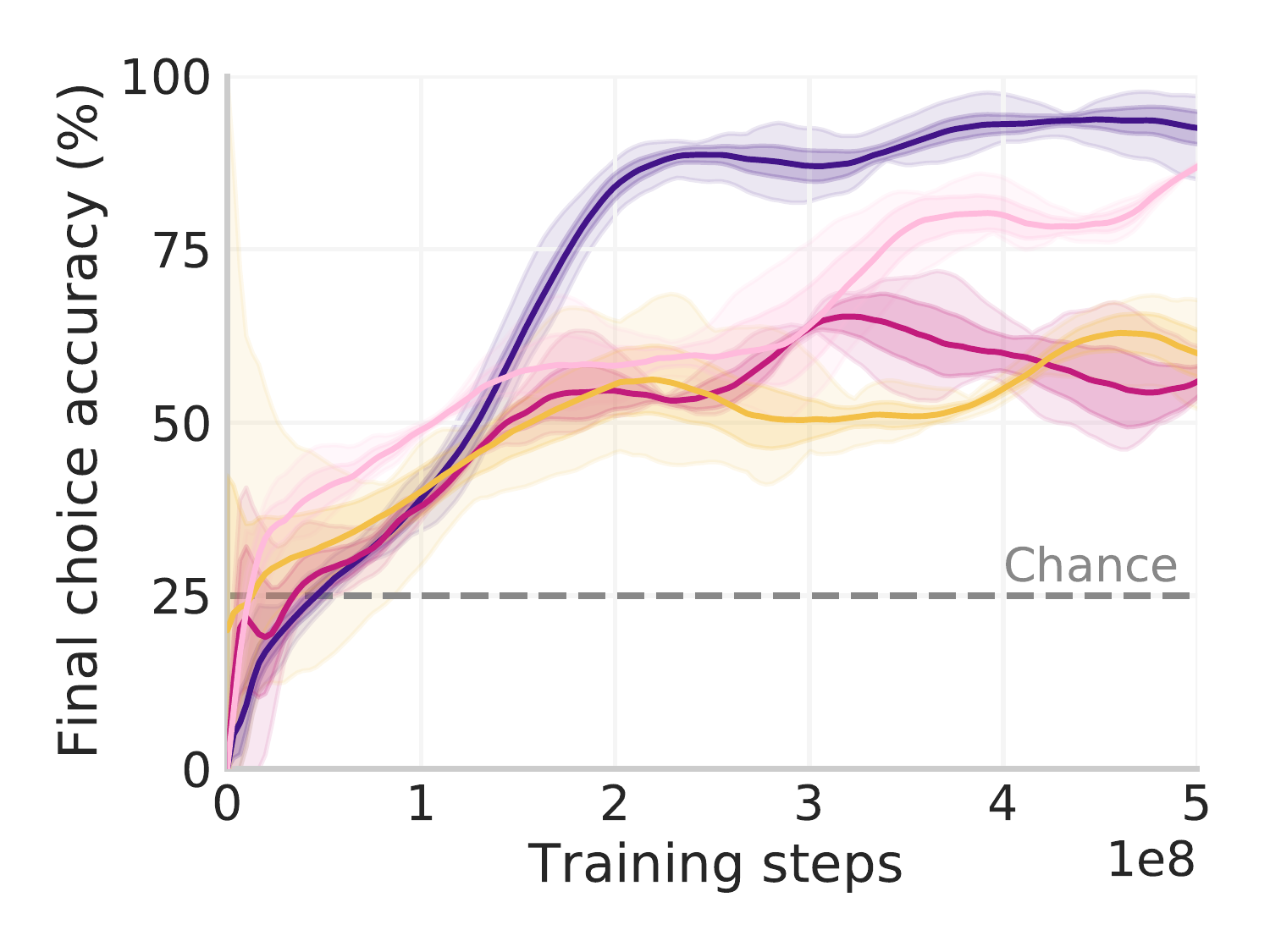}
\caption{Basic 2D.}  \label{fig:shuffled:basic}
\end{subfigure}%
\begin{subfigure}[b]{0.33\textwidth}
\centering
\includegraphics[width=\textwidth]{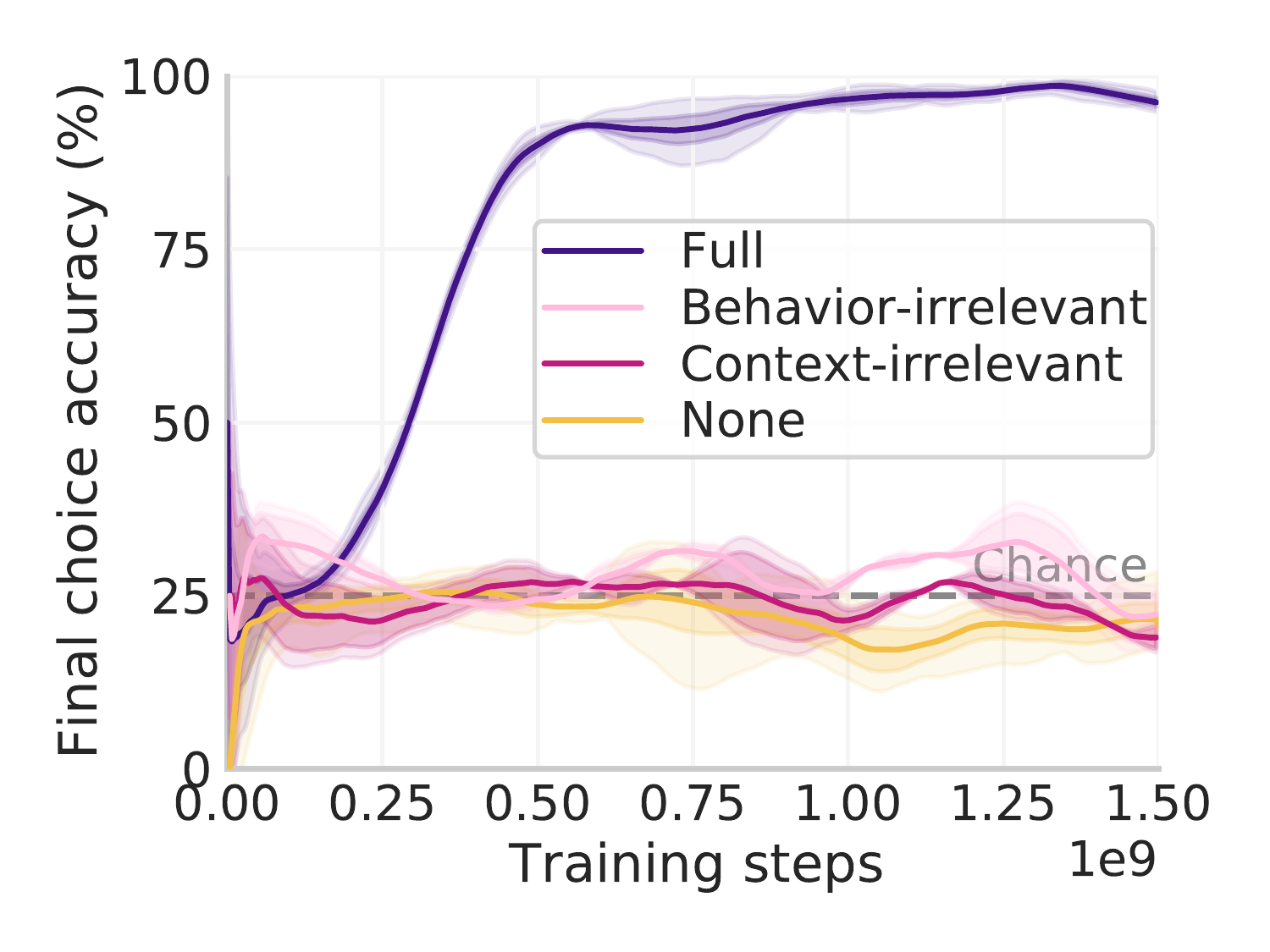}
\caption{Meta-learning (easy levels).}  \label{fig:shuffled:metaeasy}
\end{subfigure}%
\begin{subfigure}[b]{0.33\textwidth}
\centering
\includegraphics[width=\textwidth]{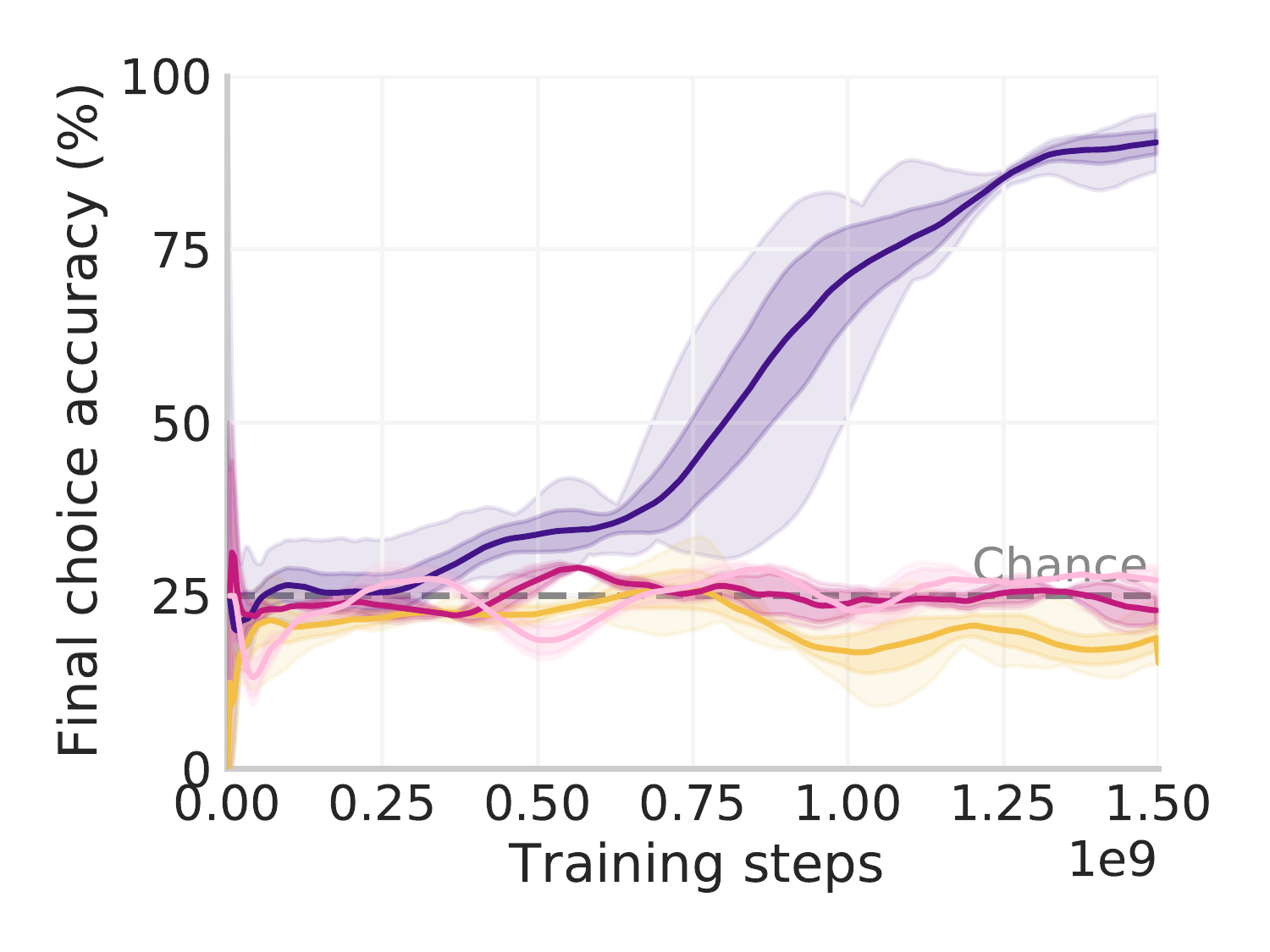}
\caption{Meta-learning (hard levels).}  \label{fig:shuffled:metahard}
\end{subfigure}%
\caption{Explanations must be behaviourally (as well as contextually) relevant to be useful in challenging settings; explanations that are contextually-irrelevant are useless in every experiment. (\subref{fig:shuffled:basic}) In the basic 2D odd-one-out tasks, behavior-irrelevant explanations eventually result in relatively comparable performance compared to full explanations, but produce much slower learning. Context irrelevant explanations are not substantially different than no explanations. (\subref{fig:shuffled:metaeasy}-\subref{fig:shuffled:metahard}) In both easy and hard learning-to-experiment levels, only an agent with full, behavior and context-relevant explanations is able to learn the tasks at all. Thus, more challenging task settings require more specific, behavior-releavnt explanations. (3 seeds for behavior-irrelevant/context-irrelevant conditions.)} \label{fig:shuffled_explanations}
\end{figure}

We next investigate whether explanations need to be relevant to the agent's behavior, or even to the situation at all, in order to be useful. To do this, we provide the agent with explanations that either are situation-relevant, but behavior irrelevant, or are irrelevant to both behavior and situational context. To produce the situation-relevant but behavior-irrelevant explanations, we first construct an episode as before. We then enumerate all the property and reward explanations that it would be possible to receive in that episode, and present a randomly selected one to the agent on approximately 10\% of steps, regardless of the agent's actions. These explanations do contain information about the objects in the scene, and can therefore potentially still benefit learning, but they do not directly react to the agent's actions.

We also considered context-irrelevant explanations that were randomly sampled from the set of all possible explanations (we chose either a property explanation or a post choice one with 50\% probability, and then sampled a random set of attributes to fill out the template). This condition is essentially a control for the possibility that predicting structured information---even meaningless information unassociated with the task---could be acting as form of regularization.

Our results (Fig. \ref{fig:shuffled_explanations}) show that explanations that are relevant to both situation and behavior are most useful, situation-relevant but behavior-irrelevant explanations can be better than nothing in some cases, and totally irrelevant explanations are not beneficial at all. Specifically, for the basic tasks behavior-irrelevant explanations still result in some learning, but are much slower than full behavior-relevant explanations. 

\subsection{Providing explanations as agent inputs is not beneficial, and interferes with learning from explanation targets} \label{app:analyses:exp_as_input}
\begin{figure}[h]
\centering
\includegraphics[width=0.4\textwidth]{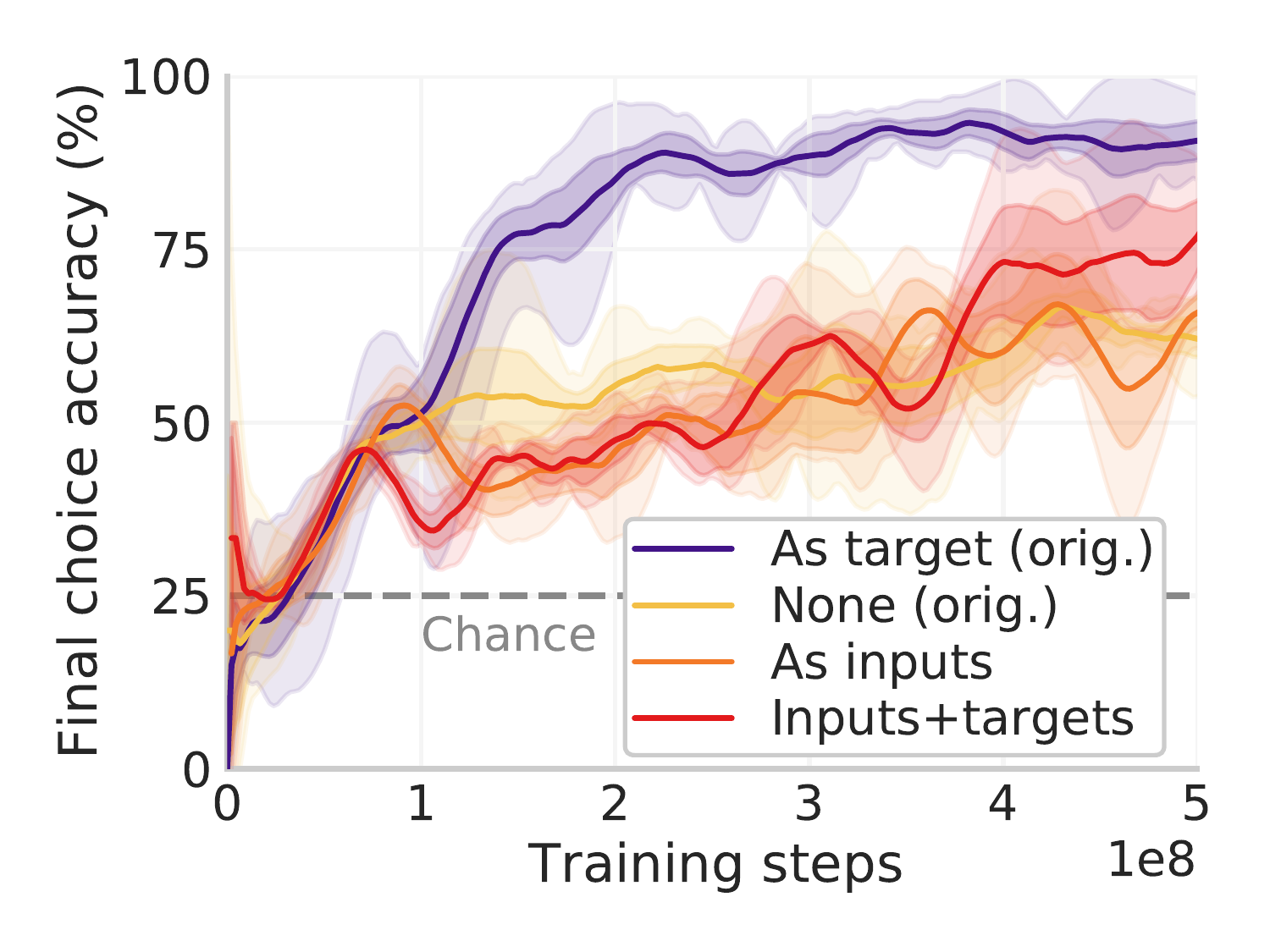}
\caption{Providing explanations as agent inputs is not beneficial (performs better than no explanations), and is actively detrimental if explanations are also used as targets. (3 seeds per as-input condition, 5 seeds for main conditions.)} \label{fig:ablations:exp_as_input}
\end{figure}

In the main text, we focused on explanations as targets during training, rather than inputs. That approach is beneficial, because it does not require explanations at test time, while explanations as input generally does. In Fig. \ref{fig:ablations:exp_as_input}, we show that furthermore \emph{providing explanations as input to the agent is not beneficial, and is actively detrimental if explanations are also used as targets}, presumably because in the latter case the agent can just ``pass through'' the explanations, without having to learn the task structure. However, it is possible that providing explanations as input on the timestep \emph{after} the agent predicts them could be useful (as in language model prediction, where each word is input after the model predicts it); we leave evaluating this possibility to future work.

\subsection{Different kinds of explanations have complementary, sometimes separable benefits} \label{app:analyses:types}

We generally provided agents with both property explanations and reward explanations. Is one of these explanations more useful than the other? Are they redundant? To answer these questions, we considered providing the agent with each kind of explanation independently. We generally find that having both types of explanations is best, and the benefits of different types depend on the setting.

In the 2D setting (Fig. \ref{fig:explanation_types:pyco}) either type of explanations alone results in learning, but both types together result in substantially faster learning. In the 3D setting (Fig. \ref{fig:explanation_types:playroom}), we find that property explanations are uniquely beneficial; perhaps because predicting explanations on encountering the objects helps the agent overcome the challenges of generating good representations for its memory.

For the meta-interventions tasks involving experimentations, we find (Fig. \ref{fig:explanation_types:meta}) that the reward explanations are uniquely beneficial, while properties explanations are not useful alone. The likely reason for this is clear when considering the episode structure---the relevance of a transformation to the final reward is much more directly conveyed by the reward explanations than the property ones. However, both types of explanations together are required for complete learning within the learning time we considered.

\subsection{The benefits of explanations depend on task complexity} \label{app:analyses:complexity}
\begin{figure}[H]
\centering
\begin{subfigure}[b]{0.33\textwidth}
\centering
\includegraphics[width=\textwidth]{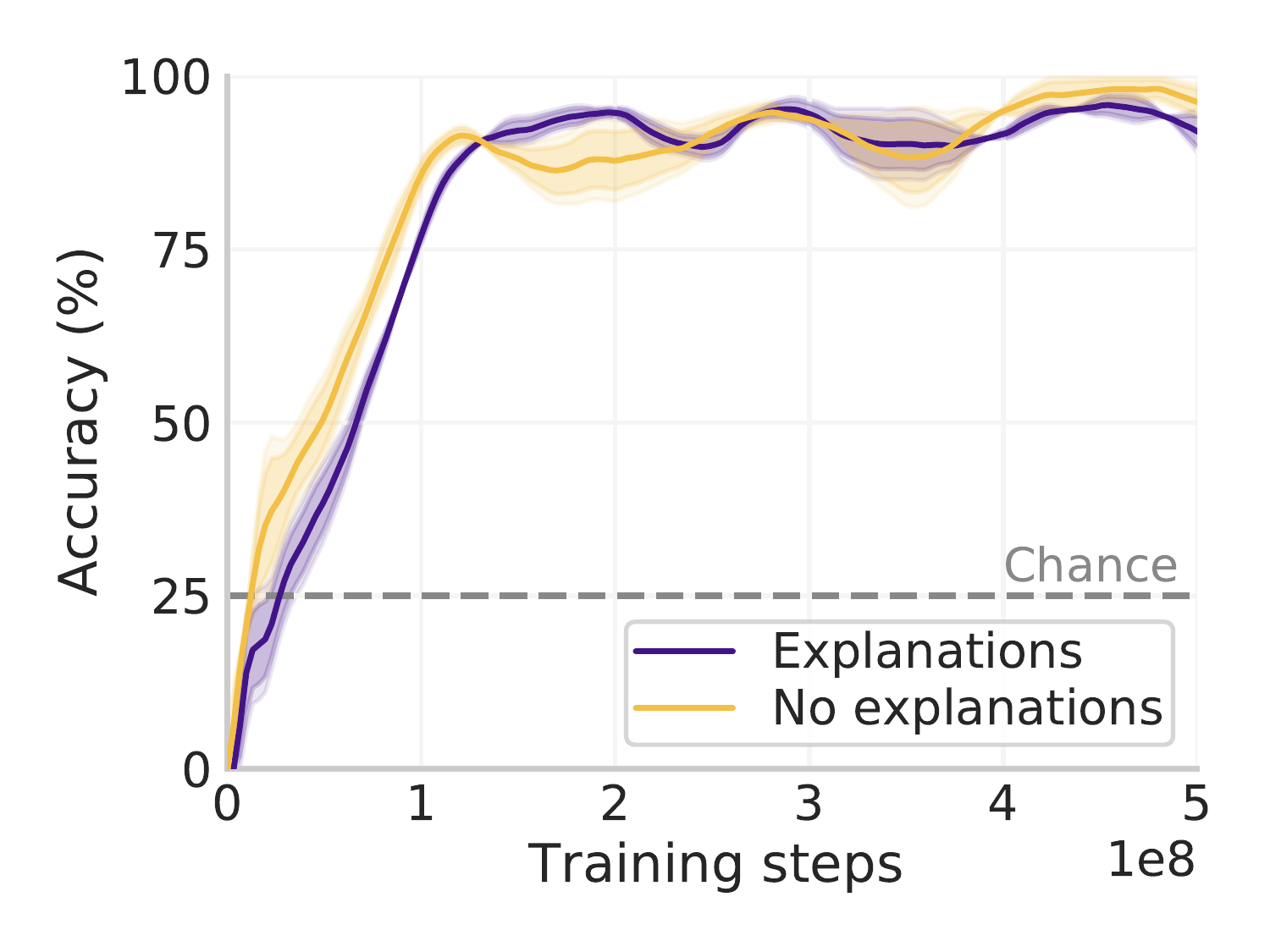}
\caption{Position \& color only.}  \label{fig:simpler:cp}
\end{subfigure}%
\begin{subfigure}[b]{0.33\textwidth}
\centering
\includegraphics[width=\textwidth]{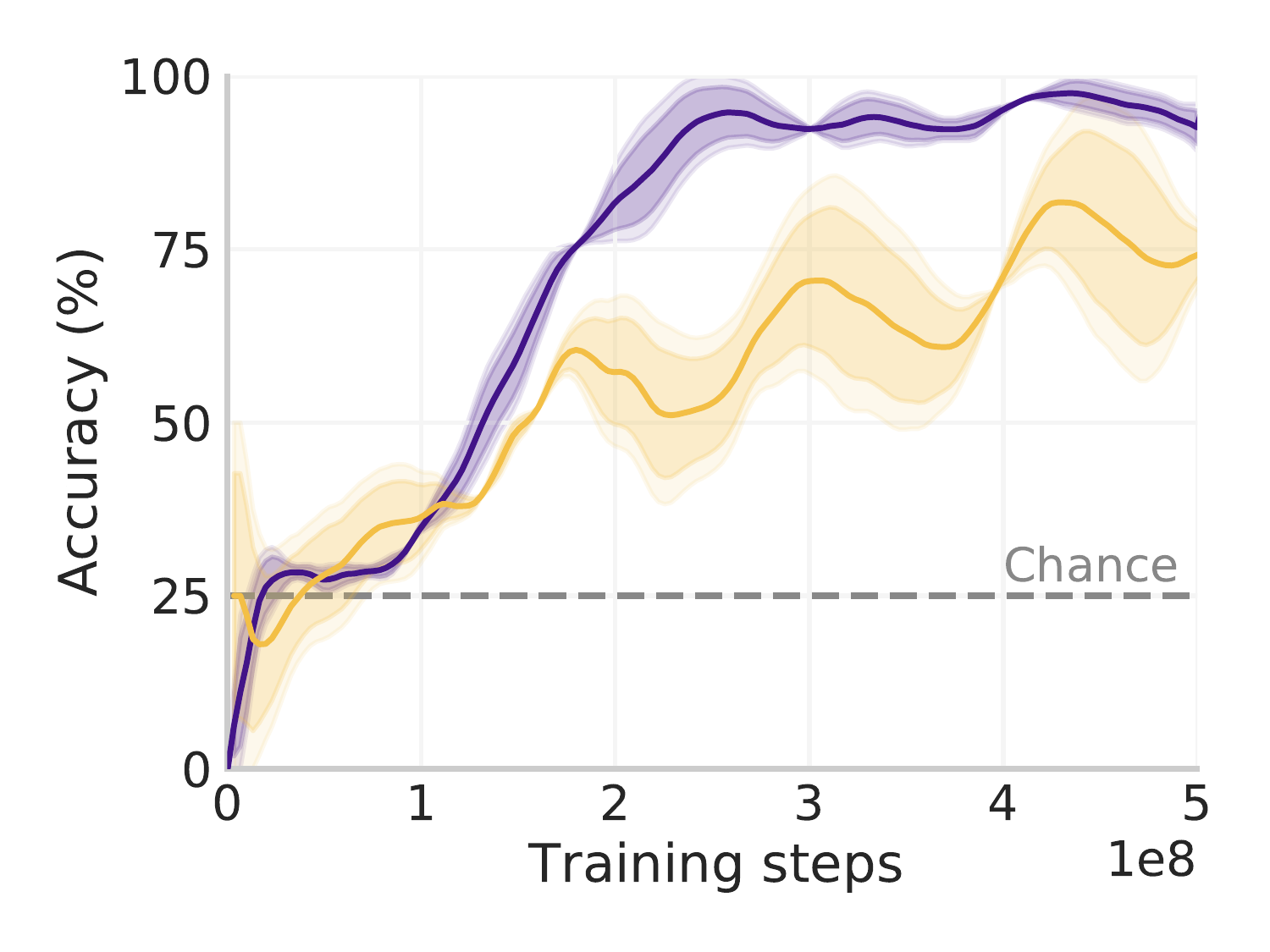}
\caption{Shape \& texture only.}  \label{fig:simpler:st}
\end{subfigure}%
\caption{The benefits of explanations depend on task difficulty. We train agents in the 2D environment on easier odd-one-out tasks where only two of the four dimensions are ever relevant---in these easier cases the agent is able to make more progress without explanations. (\subref{fig:simpler:cp}) When only the easy dimensions of position and color are ever relevant, the agents trained without explanations learn just as rapidly as the agents trained with explanations. (\subref{fig:simpler:st}) When the agents are trained on levels where only the harder dimensions of shape and texture are relevant, explanations still accelerate learning substantially. However, the agents trained without explanations achieve some learning in this condition, while they do not achieve any learning on these dimensions in the harder tasks used for the main experiments (see Fig. \ref{fig:basic_by_dimension:shape}-\subref{fig:basic_by_dimension:texture}). (2 seeds per condition.)}  \label{fig:simpler}
\end{figure}

While we generally considered tasks with many feature dimensions that might be relevant, here we show that simpler tasks in which only two dimensions vary do not always require explanations for learning. However, task complexity depends on both the number of possibly relevant dimensions and the base difficulty of those dimensions. In Fig. \ref{fig:simpler} we show that explanations are not beneficial compared to no-explanations when only the easy features of position and color are relevant. Explanations are still beneficial when the features are more difficult (shape and texture). But even in this condition, the agent without explanations exhibits some learning, while it does not learn these dimensions at all in the main experiments, where the easier dimensions are also included  (see Fig. \ref{fig:basic_by_dimension:shape}-\subref{fig:basic_by_dimension:texture}).  
Note also our results on deconfounding---in some cases explanations may help the agent to generalize in a desired way even if they are not necessary for learning the training task.

\subsection{Language prediction is learned faster than RL} \label{app:analyses:learning_dynamics} 
\begin{figure}[h]
\centering
\includegraphics[width=0.4\textwidth]{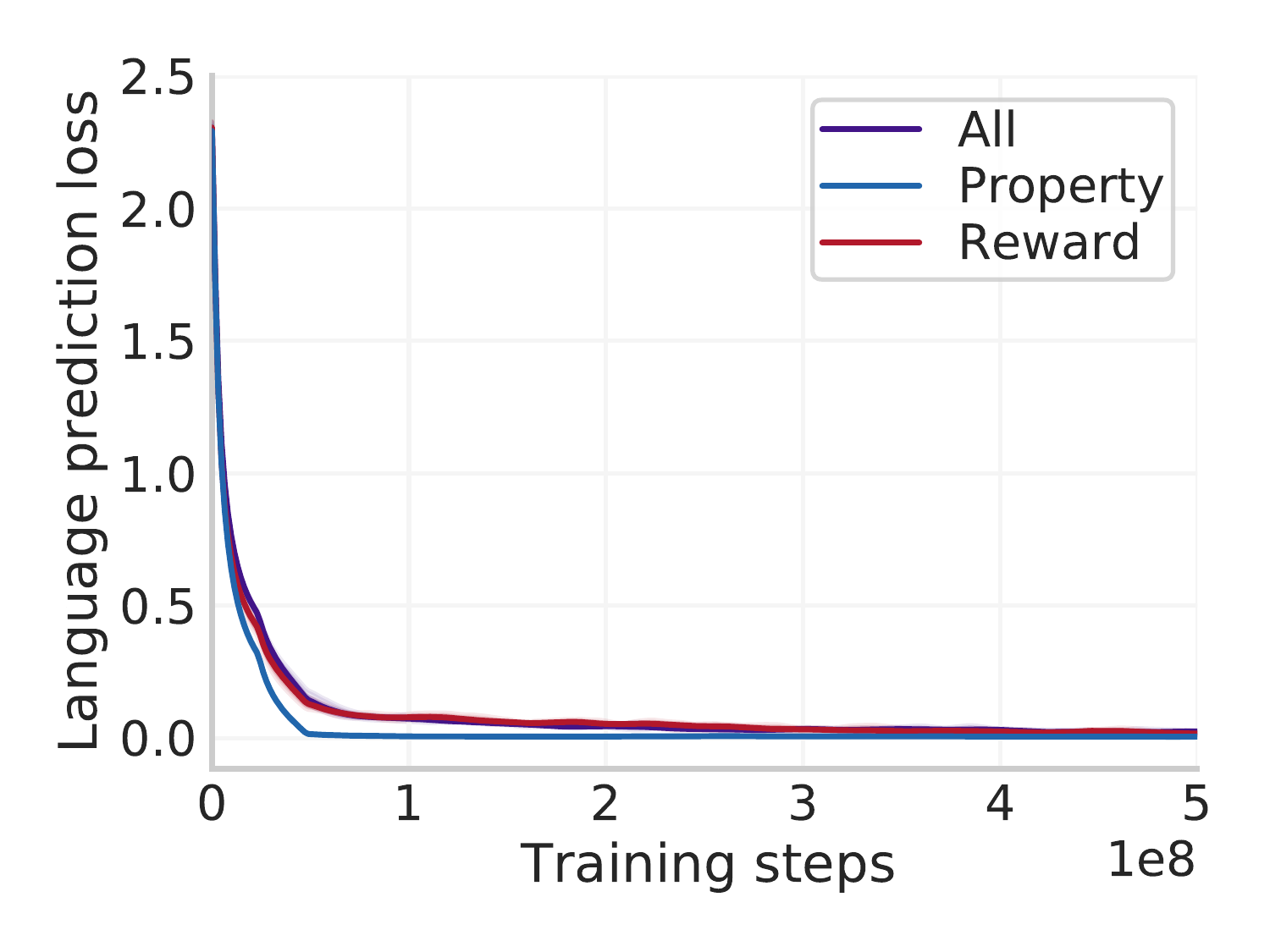}
\caption{The language prediction loss decreases rapidly early in training, relative to when the agent learns the RL tasks (evaluated on the basic 2D tasks with either full explanations, or just property or reward; compare to RL learning dynamics in Fig. \ref{fig:explanation_types:pyco}). The agent trained only with property explanations learns to predict them especially rapidly---predicting reward explanations requires more learning, unsurprisingly because these require mastering more of the task structure. However, in all cases the language prediction loss decreases substantially before the agent masters the task. (5 seeds for all explanations condition, 2 for others. Note that the lines on this plot represent different training conditions that include all or some of the explanations; we were unable to separate the language loss for the different types of predictions for the agent trained with both types.)} \label{fig:ablations:lang_loss}

\end{figure}

In Fig. \ref{fig:ablations:lang_loss} we show that the language loss decreases substantially early in training, before the agent has mastered the tasks---although the reward explanations are learned more slowly than property explanations, both are substantially learned before the agent masters the tasks. This supports our hypothesis that language, by providing a more consistent signal that emphasizes the task structure, makes this structure more learnable, and thereby supports RL. 


\subsection{Learning properties through a curriculum rather than auxiliary losses} \label{app:analyses:curriculum}

\begin{figure}[h]
\centering
\begin{subfigure}[b]{0.33\textwidth}
\centering
\includegraphics[width=\textwidth]{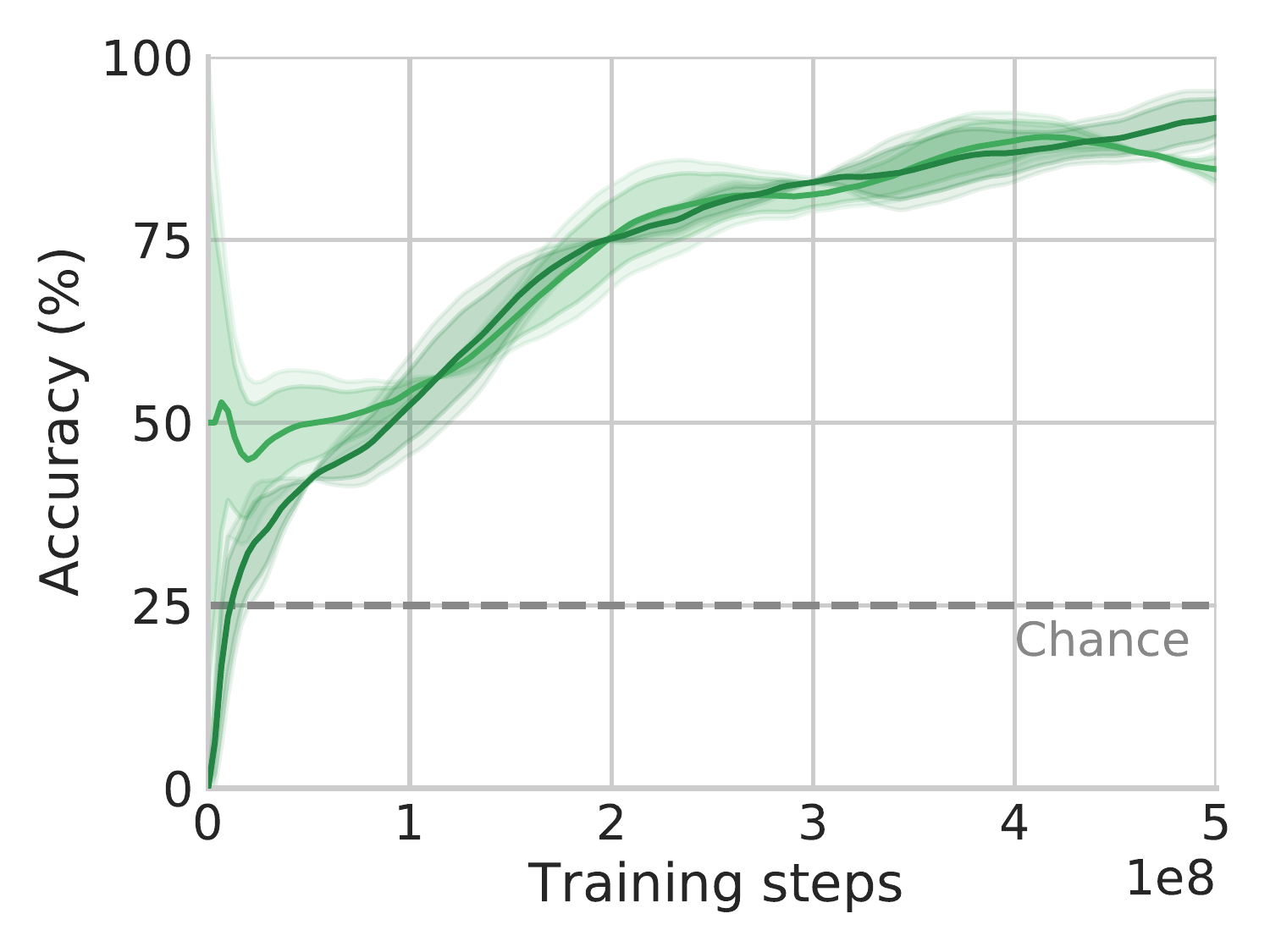}
\caption{Property-curriculum tasks.}  \label{fig:curriculum:find_tasks}
\end{subfigure}%
\begin{subfigure}[b]{0.33\textwidth}
\centering
\includegraphics[width=\textwidth]{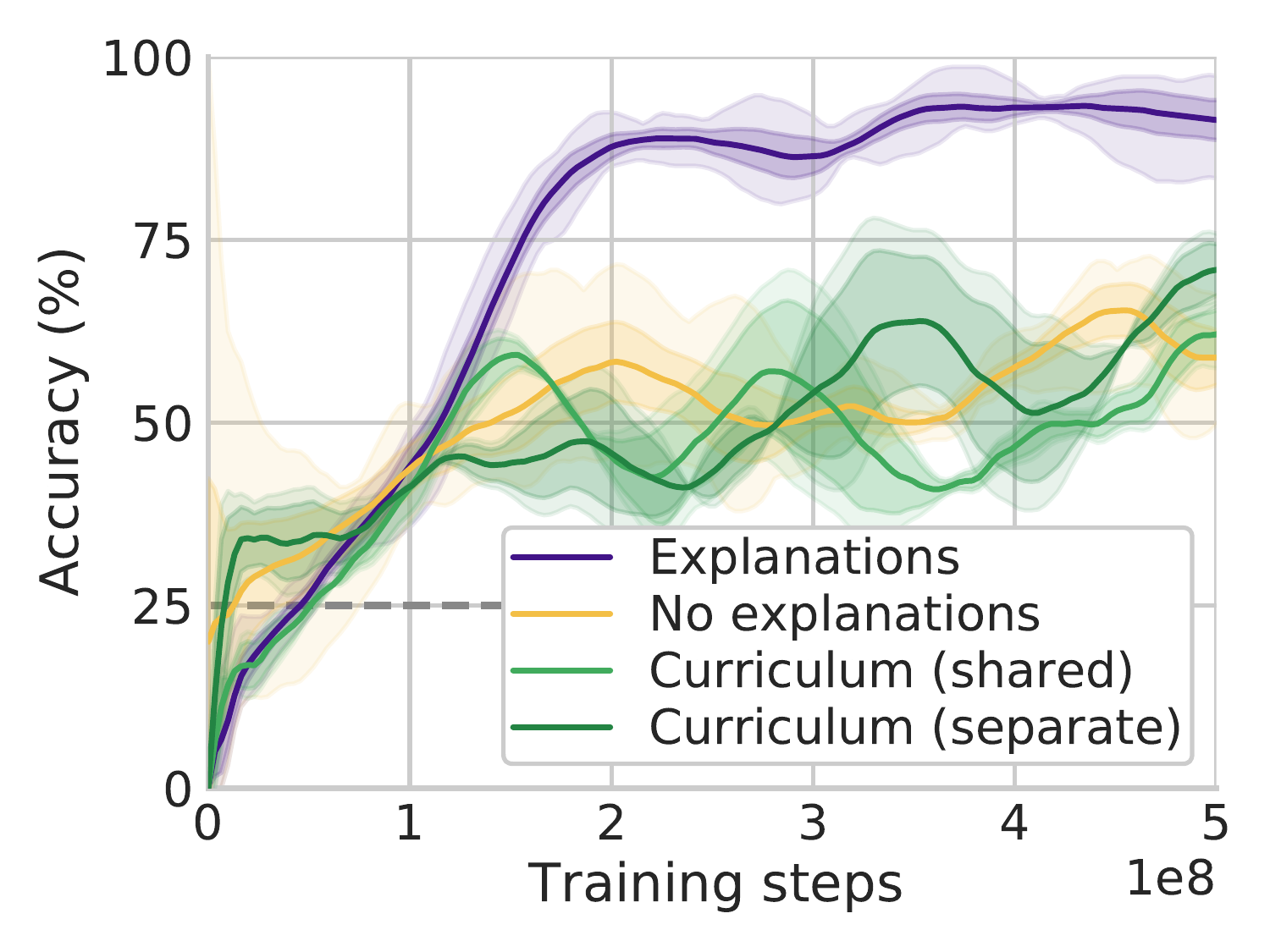}
\caption{Odd-one-out tasks.}  \label{fig:curriculum:ooo_tasks}
\end{subfigure}%
\caption{Agents trained with a curriculum of tasks that teach the properties (\subref{fig:curriculum:find_tasks}) do not learn the odd-one-out tasks (\subref{fig:curriculum:ooo_tasks}) any better than agents trained with no explanations. Results are similar whether the agent uses a shared policy (light green) for both the curriculum and odd-one-out tasks, or uses separate policies for each (dark green). (2 seeds per condition for curriculum conditions, 5 seeds for main conditions.)}  \label{fig:curriculum}
\end{figure}
Because predicting property explanations alone can be beneficial, we next consider whether the agent could benefit from learning properties through auxiliary tasks which teach those properties, rather than through explanations. Specifically, we provide the agent with a simpler property-learning task in 50\% of episodes, where it receives a property like ``red'' as an input instruction, and has to choose the corresponding object (all objects are different along each feature dimension). These tasks provide a different way to force the agent to learn the properties of the objects. On the odd-one-out tasks, the agent receives the instruction ``find the odd one out'' to distinguish its goal.

Surprisingly, we find that learning these tasks does not substantially accelerate learning of the odd-one-out tasks compared to a no-explanations and no-property-tasks baseline (Fig.\ \ref{fig:curriculum:ooo_tasks}). We initially thought this might be due to interference due to the shared policy being used for different tasks, so we reran the experiment with separate policy heads for the curriculum tasks and odd-one-out tasks, but this did not substantially change results. Auxiliary prediction of explanations may therefore be a more efficient way to encourage learning of task-relevant features, perhaps because it actively engages with the agent's behavior in the settings where those dimensions are particularly relevant.

\subsection{Auxiliary unsupervised losses are neither necessary nor sufficient; thus the benefits of explanations are not simply due to more supervision} \label{app:analyses:recon_lesion} 
\begin{figure}[h]
\centering
\includegraphics[width=0.4\textwidth]{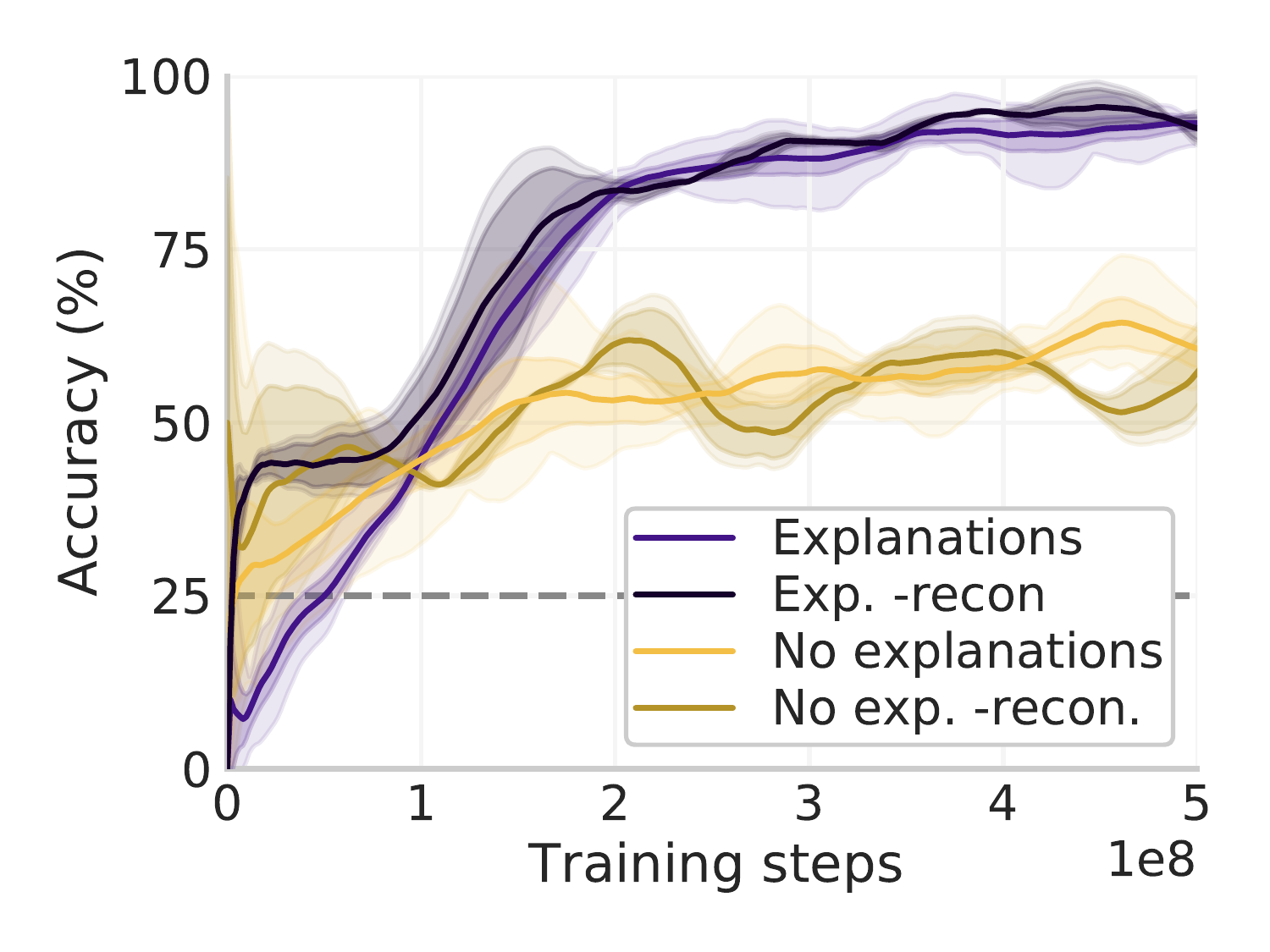}
\caption{Ablating the auxiliary reconstruction losses does not alter the pattern of results in the 2D environment---thus reconstruction is not necessary for learning these tasks. Comparing to the results which include the reconstruction losses shows that reconstruction losses are also not sufficient without explanations, even though the reconstruction losses provide substantially more supervision on every step. (2 seeds per condition.)} \label{fig:ablations:reconstruction}

\end{figure}

In Fig. \ref{fig:ablations:reconstruction} we show that the auxiliary reconstruction losses are not necessary for learning the odd-one-out tasks. Furthermore, the main text results without explanations show that these losses are not sufficient for learning either. This shows that the benefits of explanations are not simply due to having more supervision for the agent, but rather are specific to supervision that highlights the abstract task structure.

\clearpage
\section{Quantitative results} \label{app:quantitative}

\begin{table}[H]
    \centering
    \caption{Numerical results from main experiments/figures in each domain---mean \(\pm\) standard deviation across seeds. Results are average performance (\% correct) across evaluations during the last 1\% of training.} \label{tab:app:numericalresults}
    \begin{tabular}{ccccc}
      Experiment & Level & Fig. & Condition & Performance\\ \hline
      \multirow{2}{*}{Perceptual 2D} & \multirow{2}{*}{-} &  \multirow{2}{*}{\ref{fig:basic:results:pyco}} & Explanations & \(91.3 \pm 0.7 \) \\
      & & & No explanations & \(61.9\pm 2.2 \) \\ \hline
      \multirow{2}{*}{Perceptual 3D} & \multirow{2}{*}{-} &  \multirow{2}{*}{\ref{fig:basic:results:playroom}} & Explanations & \(92.7 \pm 1.4 \) \\
      & & & No explanations & \(29.5\pm 0.7 \) \\ \hline
      \multirow{6}{*}{Deconfounding} & Chose color &  \multirow{3}{*}{\ref{fig:deconfound:results:noexp}} & No explanations & \(55.4 \pm 2.6 \) \\
      & Chose shape & & No explanations & \(24.2\pm 7.6 \) \\
      & Chose texture & & No explanations & \(15.4\pm 6.7 \) \\
      & Chose color & \multirow{3}{*}{\ref{fig:deconfound:results:explanations}} & Explain color & \(95.5\pm 0.9 \) \\
      & Chose shape & & Explain shape & \(87.5\pm 2.9 \) \\
      & Chose texture & & Explain texture & \(86.2\pm 0.9 \) \\ \hline      
      \multirow{4}{*}{Meta-learning} & \multirow{2}{*}{Easy} &  \multirow{2}{*}{\ref{fig:interventions:results:easy}} & Explanations & \(96.9 \pm 0.3 \) \\
      & & & No explanations & \(24.0\pm 0.6 \) \\
      & \multirow{2}{*}{Hard} &  \multirow{2}{*}{\ref{fig:interventions:results:hard}} & Explanations & \(90.5 \pm 1.6 \) \\
      & & & No explanations & \(24.6 \pm 1.2 \) \\ \hline
    \end{tabular}
\end{table}
\clearpage

\section{Methods} \label{app:methods}

\subsection{RL agents \& training} \label{app:methods:RL}

\begin{table}[H]
    \centering
    \caption{Hyperparameters used in main experiments. Where only one value is listed across both columns, it applies to both.}
    \resizebox{\textwidth}{!}{
    \begin{tabular}{|c|c|c|}
      \hline
      & \textbf{2D} & \textbf{3D} \\ \hline
      \hline
      All activation fns & \multicolumn{2}{c|}{ReLU} \\ \hline
      State dimension & \multicolumn{2}{c|}{512} \\ \hline
      Memory dimension & \multicolumn{2}{c|}{512} \\ \hline
      Memory layers & \multicolumn{2}{c|}{4} \\ \hline
      Memory num. heads & \multicolumn{2}{c|}{8} \\ \hline
      TrXL extra length & \multicolumn{2}{c|}{128} \\ \hline
      \hline
      Visual encoder & CNN & ResNet\\ \hline 
      Vis. enc. channels & \multicolumn{2}{c|}{(16, 32, 32)} \\ \hline 
      Vis. enc. filt. size & (9, 3, 3) & (3, 3, 3)\\ \hline 
      Vis. enc. filt. stride & (9, 1, 1) & (2, 2, 2)\\ \hline 
      Vis. enc. num. blocks & NA & (2, 2, 2)\\ \hline 
      \hline
      Policy \& value nets & \multicolumn{2}{c|}{MLP with 1 hidden layer with 512 units.} \\ \hline
      Reconstruction decoder & \multicolumn{2}{c|}{Architectural transpose of the encoder, with independent weights.} \\ \hline
      Explanation decoder & \multicolumn{2}{c|}{1-layer LSTM} \\ \hline
      Explanation LSTM dimension & \multicolumn{2}{c|}{256} \\ \hline
      \hline
      Recon. loss weight & 1. & 0. \\ \hline
      \(V\)-trace loss weight & \multicolumn{2}{c|}{1.} \\ \hline
      \(V\)-trace baseline weight & \multicolumn{2}{c|}{0.5} \\ \hline
      Explanation loss weight & \(3.3 \cdot 10^{-2}\) for main, \(0.2\) for meta-learning. & \(3.3 \cdot 10^{-2}\)\\ \hline
      Entropy weight & \(1\cdot10^{-2}\) & \(1\cdot10^{-3}\) \\ \hline
      \hline
      Batch size & \multicolumn{2}{c|}{24} \\ \hline
      Training trajectory length & \multicolumn{2}{c|}{50} \\ \hline
      \hline 
      Optimizer & \multicolumn{2}{c|}{Adam \citep{kingma2014adam}} \\ \hline
      LR & \(1\cdot10^{-4}\) & \(5\cdot10^{-5}\) with explanations, \(2\cdot10^{-5}\) without\\ \hline
    \end{tabular}}
    \label{tab:app:hyper}
\end{table}

In Table \ref{tab:app:hyper} we list the architectural and hyperparameters used for the main experiments. All agents were implemented using JAX \citep{jax2018github} and Haiku \citep{haiku2020github}, and were trained using TPU v3 and v4 devices.

\paragraph{Additional architectural details:} The agent visual encoder output is flattened and then linearly projected to a single vector; the transformer memory receives this one vector input per timestep, and the corresponding single transformer memory output vector for the current timestep is used as input to the heads. The transformer memory uses relative position encodings. The visual decoders use depth-to-space upsampling where necessary, i.e. they create a feature map of shape \(N \times M \times (C \cdot k^2)\) and then reshape by location to get a result of shape \((N \cdot k) \times (M \cdot k) \times C\). The agent outputs a full language prediction per timestep

\paragraph{Hyperparameter choices and tuning:} In most cases the hyperparameters were taken from other sources without tuning for our setup. There are two main exceptions: (1) we chose the explanation weight loss to approximately balance the magnitude of this loss with the magnitudes of the RL losses early in training in each experiment, and (2) we swept the learning rate for the main experiments in 2D and 3D (but used the same settings for follow-up experiments, including meta-learning). In the 2D experiments we found that a similar learning rate was best for both agents trained with and without explanations, but in 3D we found that agents trained without explanations needed a slower learning rate to avoid their performance degrading from chance-level to below chance. 

Since we ran the 3D experiments after 2D, we used similar hyperparameters, except that we found we needed to decrease the learning rate as noted above. However, some hyperparameters do differ across tasks due to specific task features. For example, the visual encoder for the 2D tasks is set to have a filter size of 9 because this is the resolution of each square in the grid, and the entropy cost for 2D tasks was chosen from prior work which used a similar grid world action space \citep{hill2019environmental}, while the 3D cost is lower because of the more complex action space for these tasks (see below). These decisions were shared across experimental conditions, and the choices were based upon prior work in similar environments, so should not favor one condition over another.

\paragraph{Explanation prediction loss:} We trained the agents to predict the language explanation using a softmax cross-entropy loss over a word-level vocabulary of 1000 tokens (more than were necessary for the limited language we used). The explanation loss was summed across the sequence of tokens. 

\paragraph{Self-supervised image reconstruction loss:} We trained the agents to reconstruct the image pixels (normalized to range [0, 1] on each color channel) with a sigmoid cross-entropy loss. The image reconstruction loss was averaged across all pixels and channels. However, we found in follow-up experiments that this did not substantially change results in 2D (Appx. \ref{app:analyses:recon_lesion}), so we disabled this loss in 3D.

\subsection{RL environment details} \label{app:methods:environments}

We are in the process of preparing our 2D environments for release, and will update this paper once we have done so.

\subsubsection{2D}
The 2D tasks were implemented in Pycolab (\url{https://github.com/deepmind/pycolab}). instantiated in a \(9 \times 9\) tile room with an extra 1 tile wall surrounding on all sides, for a total of \(11 \times 11\) tiles. This was upsampled at a resolution of 9 pixels per tile to form a \(99 \times 99\) image as input to the agent. The agent was placed in the center of the room, and had 9 possible actions, allowing it to move one square in any of the 8 possible directions, or to do nothing.

Four objects were placed in the room with the agent. They were chosen so that a single object was the odd one out, along a single dimension, and features appeared in two pairs along the other dimensions. The objects varied along the feature dimensions of:
\begin{itemize}
    \item Color: one of 19 possible colors (e.g. green or lavender).
    \item Shape: one of 11 possible shapes (e.g. triangle or tee).
    \item Texture: one of 6 possible textures (e.g. horizontal stripes or checkers).
    \item Position: One of 4 position types (in corner, against horizontal wall, against vertical wall, or in a 3x3 square in the center). 
\end{itemize}

The agent was given 128 steps to complete each episode, after which the episode would immediately terminate. The agent had to choose an object by walking onto the grid cell containing it. It would be immediately rewarded 1 if the object was the odd one out, and 0 otherwise. However, the episode would last for an additional few steps to give the agent time to learn from the reward explanation (if provided); this extra time was provided even if the agent was not trained to predict explanations, in order to precisely match the training experience across conditions. No additional reward would be received during this period, but the agent would be asked to output the reward explanation at every timestep. This period would last either 16 steps, until the agent touched any object, or until the full episode limit of 128 steps was reached, whichever came soonest. (However, this full extra length was likely unnecessary, see below.)

If the agent was trained to predict property explanations, whenever it was adjacent to an object it would be asked to predict a string of the form:
\begin{verbatim}
    This is a red horizontal-striped triangle in-the-corner
\end{verbatim}
The properties always appeared in the order color texture shape position. These sentences were tokenized at a word level, with the hyphenated phrases treated as single words. Hence, a single token was attached to each possible feature value along each dimension.

If the agent was adjacent to multiple objects, which description it received was determined randomly. Once the agent made a choice, property descriptions were disabled.

If the agent was asked to predict reward explanations, for the period after receiving the reward (see above), it would be asked to predict a string in one of the following forms:
\begin{verbatim}
    Correct because it is uniquely horizontal-striped
    
    Incorrect because other objects are red horizontal-striped
    triangles or in-the-corner
\end{verbatim}
Thus, the reward explanations identify all features that contributed to a decision being incorrect. 

\textbf{Does the agent need to predict the reward explanation for 16 steps?} We gave the agent 16 steps to predict the reward explanations to make the signal more salient. However, in a follow-up experiment (not shown) where the agent had only a single step for prediction in 2D domain, we observed only a very minor slowdown in learning, even if the agent was only given reward explanations. Thus we do not believe that this extra time is strictly necessary (although we have not thoroughly explored this across all experimental domains).

\textbf{Confounding:} The confounded training tasks were performed in a variation of the 2D setup. All objects were initialized to have the same position type (as defined above; randomly selected), but distinct positions. One object was chosen to be the target. The other three objects were all assigned matching colors, shapes, and textures, while the target object was assigned a unique color, shape, and texture.

When explanations were provided, both property and reward explanations described only a single feature dimension in any given condition.

\textbf{Deconfounded tests:} The deconfounded tests, used for both the confounding experiments and the final trial of the meta-learning experiments (below), assigned all objects to the same position type. A common color, shape, and texture were chosen, as well as rare versions of each. One object was assigned common versions of all three attributes, the second was assigned the rare color but common shape and texture, the third was assigned the rare shape but common color and texture, and the final one was assigned the rare texture but common shape and color. 

\textbf{Meta-learning to experiment:} For the experimentation tasks, the agent received a series of four trials. The first three trials were experimentation trials, while the fourth was a deconfounded test (as above), except that on the final trial the agent received a reward of 10 for a correct answer instead of 1. The agent was given 512 steps to complete all 4 trials---i.e., average time allotted per trial equal to the tasks above.

For the experimentation trials, the agent was granted three additional actions, which would transform the corresponding attributes (color, shape, or texture) of an adjacent object. If the agent was adjacent to multiple objects, the target was chosen pseudo-randomly (according to the order in the environment's internal object representation). The agent could only employ one transformation action per trial. After the first time, these actions did not have any effect. If the agent was not adjacent to an object, or during the deconfounded test trial, the actions also had no effect. 

The objects for the experimentation trials were again initialized with matching position types. The other attributes could either all be matching (easiest levels), or could be paired (harder levels)---in the latter case, there would be two values for that attribute that each appeared on two objects, e.g., two red objects and two blue. The results shown in the main figures are from levels where all are matching, or all are paired, but we also trained the agent on intermediate levels where some attributes were paired, but some were matching, which helped the agent to transition from learning the easy levels to the hardest ones.

The reward explanations that agents received named which dimension was relevant in the current episode. Neither property descriptions nor reward explanations mentioned position, as it was never a relevant concept.

In order for the agent to determine which type of trial it was performing, it was allowed to observe an input instruction that either said ``Make an odd one out'' (experimentation trials) or ``Find the odd one out'' (final trial). This was provided through a separate channel from the explanations; as before, the agent did not observe the explanations.

\subsubsection{3D}

The 3D environments were implemented in Unity.
The agent was placed in a room with a randomly-located door and windows (the agent could not interact with these). It had 10 possible actions: moving forward or backard, moving left or right, looking left or right or up or down, grabbing an object it was facing (if within a certain distance), and doing nothing.

Four objects were placed in the room, with attributes sampled as in the 2D environment from the dimensions:
\begin{itemize}
    \item Color: one of 10 possible colors (e.g. blue or magenta).
    \item Size: one of 3 possible sizes (small, medium, or large).
    \item Texture: one of 6 possible textures/materials (e.g. metallic or wood-grain).
    \item Position: One of 3 position types (in corner, against wall, or in the center). 
\end{itemize}

The episode lasted for 60 seconds, at 30 FPS; but the agent took an action only once every 4 frames (the action was then repeated until the next agent step), so the episode lasted for at most 450 agent steps. The agent had to use its ``grab'' action on an object to make a choice; colliding with the objects would simply cause them to move. 25 steps were allocated for reward explanations (if any).

If the agent was asked to predict property explanations, they were given when the agent was facing an object and close enough to grab it. 

The property and reward explanations for the 3D environment were analogous to the ones for the 2D environment, except that the reward explanations did not have the ``or'' before the final attribute.


\end{document}